\documentclass[12pt]{article}

\usepackage{newtxtext,newtxmath}

\usepackage{graphicx}

\usepackage[letterpaper,margin=1in]{geometry}

\renewenvironment{abstract}
	{\quotation}
	{\endquotation}

\date{}

\makeatletter
\renewcommand{\fnum@figure}{\textbf{Figure \thefigure}}
\renewcommand{\fnum@table}{\textbf{Table \thetable}}
\makeatother

\usepackage{scicite}

\usepackage{url}
\usepackage{multicol}
\usepackage{color}
\usepackage{colortbl}  

\usepackage{array} 
\usepackage{float}
\usepackage{chemformula}
\usepackage{url}
\usepackage{algorithm}  
\usepackage{algorithmicx} 
\usepackage{algpseudocode}
\usepackage{adjustbox}

\usepackage{ragged2e}
\usepackage{wrapfig}
\usepackage{pifont}
\usepackage[accsupp]{axessibility}
\makeatletter
\let\NAT@parse\undefined
\makeatother
\usepackage[numbers,sort&compress]{natbib}

\usepackage{marvosym}

\usepackage{booktabs}                                   
\usepackage{multirow}                                   
\usepackage{makecell}                                   
\usepackage{tablefootnote}                              
\usepackage[symbol]{footmisc}                           
\usepackage{amsmath,amssymb}                            
\usepackage{xcolor}                                     
\usepackage{enumitem}                                   
\usepackage{subcaption}                                 
\usepackage{stfloats}                                   

\usepackage[bookmarks=true]{hyperref}                   
\usepackage{graphicx} 

\usepackage{epsfig}
\usepackage{epstopdf} 

\usepackage{cleveref}
\usepackage{csquotes}
\usepackage{xspace}

\usepackage{nicematrix}
\usepackage{bm}
\usepackage{pifont}%
\newcommand{\cmark}{\ding{51}}%
\newcommand{\xmark}{\ding{55}}%

\newcommand{\dgray}[1]{\textcolor{darkgray}{#1}}
\newcommand{\gray}[1]{\textcolor{gray}{#1}}
\newcommand{\orange}[1]{\textcolor{orange}{#1}}

\newcommand{\eat}[1]{}                                  

\newcommand{\ours}[0]{{RoboMIND~2.0}\xspace}
\newcommand{\ntrajs}{310K\xspace}
\newcommand{\ntasks}{759\xspace}
\newcommand{\nobjs}{1139\xspace}
\newcommand{\nskills}{129\xspace}

\def\scititle{
	RoboMIND 2.0: A Multimodal, Bimanual Mobile Manipulation Dataset for Generalizable Embodied Intelligence
}
\title{\bfseries \boldmath \scititle}

\author{%
 Chengkai Hou$^{1,2,*,\ddagger}$, Kun Wu$^{1,*,\ddagger}$, Jiaming Liu$^{2,*,\ddagger}$, Zhengping Che$^{1,*,\dagger}$, Di Wu$^{1,2,*}$,\\
 Fei Liao$^{1,*}$, Guangrun Li$^{1,2,*}$, Jingyang He $^{1,2,*}$, Qiuxuan Feng$^{1,2,*}$, Zhao Jin$^{1,*}$,\\
 Chenyang Gu$^{2}$, Zhuoyang Liu$^{2}$, Nuowei Han$^{2}$, Xiangju Mi$^{2}$, Yaoxu Lv$^{2}$,\\
 Yankai Fu$^{2}$, Gaole Dai$^{2}$, Langzhe Gu$^{2}$, Tao Li$^{1}$, Yuheng Zhang$^{1}$, Yixue Zhang$^{1}$, \\
 Xinhua Wang$^{1}$,
 Shichao Fan$^{1}$, Meng Li$^{1}$, Zhen Zhao$^{1}$, Ning Liu$^{1}$,\\
 Zhiyuan Xu$^{1}$, Pei Ren$^{1}$, Junjie Ji$^{1}$, Haonan Liu$^{1}$,\\
 Kuan Cheng$^{2}$,
 Shanghang Zhang$^{2,\text{\Letter}}$, Jian Tang$^{1,\text{\Letter}}$\\
 $^1$Beijing Innovation Center of Humanoid Robotics\\
 $^2$State Key Laboratory of Multimedia Information Processing, \\School of Computer Science, Peking University\\
}

\begin{document} 
\footnotetext[0]{%
 \\
 $^{1}$Beijing Innovation Center of Humanoid Robotics, Beijing, China. Kun Wu: \textit{Gongda.Wu@x-humanoid.com}; Zhengping Che: \textit{z.che@x-humanoid.com}; Jian Tang:
 \textit{jian.tang@x-humanoid.com}%
 \\ 
 $^{2}$State Key Laboratory of Multimedia Information Processing, School of Computer Science, Peking University, Beijing, China. Chengkai Hou:
 \textit{2501111947@stu.pku.edu.cn}; Jiaming Liu: \textit{jiamingliu@stu.pku.edu.cn};
 Shanghang Zhang: \textit{shanghang@pku.edu.cn}%
 \\
 $^{*}$Co-first authors: Chengkai Hou, Kun Wu, Jiaming Liu, Di Wu, Fei Liao, Guangrun Li, Jingyang He, Qiuxuan Feng, Zhao Jin, and Zhengping Che; 
 $^{\ddagger}$Co-first core authors: Chengkai Hou, Kun Wu, and Jiaming Liu;
 $^{\dagger}$Project lead: Zhengping Che; 
 Other co-first authors are listed in alphabetical order;
 $^{\text{\Letter}}$Corresponding authors: Shanghang Zhang and Jian Tang.
}

\maketitle

\maketitle
\begin{center}
  \centering
  \begin{minipage}[t]{\linewidth}
    \includegraphics[width=0.99\textwidth]{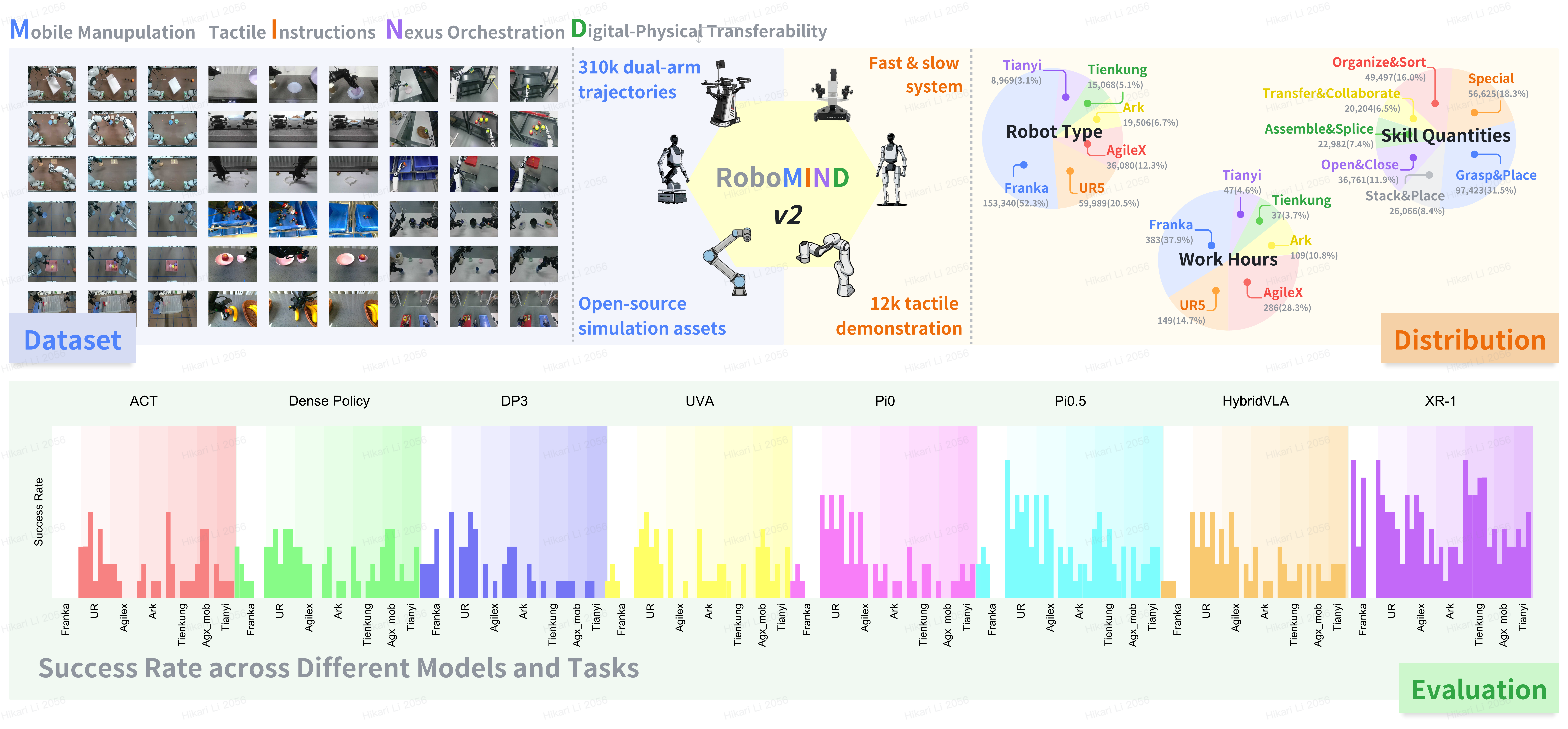}
    {\captionsetup{hypcap=false}  
    \captionof{figure}{\footnotesize{\textbf{
    Overview of the \ours.} We introduce \ours, a large-scale dataset comprising 310K dual-arm trajectories collected from six heterogeneous robot embodiments, totaling over 1,000 hours. The dataset contains rich modalities, including 12K tactile-enriched sequences and 20K mobile manipulation trajectories. Collected through a unified teleoperation and quality assurance pipeline, \ours ensures consistent proprioception and provides fine-grained natural language annotations. To support scalable training and evaluation, we release digital-twin USD assets and 20K simulation trajectories aligned with real-world tasks. Building on this foundation, we propose MIND-2, a dual-system controller that integrates a slow high-level planner MIND-2-VLM with a fast low-level policy MIND-2-VLA, enabling robust long-horizon mobile manipulation across diverse scenarios. We assess \ours across four single-task imitation learning methods (ACT, Dense Policy, DP3, and UVA) and four multi-task VLA models ($\pi_{0}$, $\pi_{0.5}$, HybridVLA, and XR-1).
    }}
    \label{fig:teaser}}
  \end{minipage}
\end{center}

\begin{abstract} \bfseries \boldmath

Data-driven imitation learning has revolutionized robotic manipulation, but existing approaches remain limited by the scarcity of large-scale, diverse real-world demonstration data, leading to insufficient generalization in long-horizon bimanual tasks and mobile manipulation in strange environments.
To bridge this gap, we present RoboMIND 2.0, a comprehensive real-world dataset comprising over $310\text{K}$ dual-arm manipulation trajectories collected across six distinct robot embodiments and $739$ complex tasks. 
Crucially, to support research in contact-rich and spatially extended tasks, the dataset incorporates $12\text{K}$ tactile-enhanced episodes and $20\text{K}$ mobile manipulation trajectories. 
Complementing these physical data, we construct high-fidelity digital twins of our real-world environments, releasing an additional $20\text{K}$-trajectory simulated dataset to facilitate robust sim-to-real transfer. 
To fully exploit the potential of RoboMIND~2.0, we propose \textbf{MIND-2} system, a hierarchical dual-system framework optimized via offline reinforcement learning. 
MIND-2 integrates a high-level semantic planner \textbf{(MIND-2-VLM)} to decompose abstract natural language instructions into grounded subgoals, coupled with a low-level Vision-Language-Action executor \textbf{(MIND-2-VLA)}, which generates precise, proprioception-aware motor actions. 
Extensive evaluations across six distinct robotic embodiments validate the effectiveness of our dataset and demonstrate that the \textbf{MIND-2} system significantly outperforms four single-task baselines covering both 2D images and 3D point clouds as well as four state-of-the-art VLA models.
Furthermore, we observe that integrating tactile modalities yields measurable gains in fine-grained manipulation tasks. 
Finally, experimental results show that mixing real and simulated data during training consistently enhances physical execution performance, validating both the fidelity of our simulation benchmarks and the cost-efficiency of synthetic data augmentation. 
Our full dataset is publicly released on \url{https://www.modelscope.cn/collections/X-Humanoid/RoboMIND20}.

\end{abstract}

\noindent

\section{Introduction}

Manipulation is a core challenge in robotics, and attaining human-level proficiency in dynamic and dexterous manipulation tasks has long been a central goal of the field~\cite{kim2024openvla,black2024pi_0,black2025pi_0.5,liu2024rdt}.
High-quality datasets hold the promise of enabling data-driven acquisition of complex and dexterous robotic skills~\cite{o2024open,khazatsky2024droid,wu2024robomind}. By learning from rich, diverse, and task-relevant demonstrations, an effective imitation  learning method can acquire highly proficient manipulation policies that are well-aligned with the physical constraints and dynamics of real-world deployment scenarios~\cite{su2025dense,ze20243d,chi2023diffusion,gervet2023act3d,kim2024openvla,liu2024rdt,liu2025hybridvla,black2024pi_0,black2025pi_0.5,fan2025xr}. Just as human expertise is built through repeated observation and practice, robots rely on large-scale, high-fidelity datasets to generalize across tasks, objects, and environments, which makes dataset quality and coverage a cornerstone of scalable robot learning.

While many publicly available robotic manipulation datasets claim to offer diversity, this diversity is typically confined to a single dimension, such as object variety, task types, environments, or robot embodiments, rather than providing comprehensive, multi-dimensional coverage essential for robust and generalizable policy learning. For instance, widely used benchmarks like Open~X-Embodiment~\cite{o2023open}, RH20T~\cite{fang2024rh20t}, DROID~\cite{khazatsky2024droid}, and RoboMIND 1.0~\cite{wu2024robomind} primarily consist of single-arm, fixed-base manipulation data and lack any examples of bimanual coordination, despite its prevalence in real-world scenarios.
Recent efforts have begun to address this gap. Datasets such as AgiBot World~\cite{bu2025agibot} and Galaxea Open-World~\cite{jiang2025galaxea} introduce rich bimanual manipulation data with diverse tasks and realistic execution contexts, and notably include high-resolution tactile sensing, which is an important step toward multimodal perception. However, both rely exclusively on a single robot embodiment (typically a humanoid platform), which severely limits their utility for studying cross-embodiment generalization. On the other hand, RoboCOIN~\cite{wu2025robocoin} expands embodiment diversity by collecting data across multiple dual-arm platforms and provides a digital twin for simulation. Yet, the per-embodiment task coverage remains sparse, with limited trajectories and scenarios per robot, hindering effective training of long-horizon, temporally extended policies.
Compounding these limitations, the vast majority of existing datasets, including even the most recent ones, capture only visual observations and basic actuation states, omitting critical physical interaction signals such as tactile feedback. This absence not only reduces multimodal richness but also impairs a model’s ability to reason about contact, slip, and fine manipulation—capabilities central to dexterous robotic behavior.

To contextualize these gaps,  we introduce \ours, a comprehensive dataset of robotic manipulation comprising approximately \ntrajs bi-manual operation trajectories from \ntasks manipulation tasks collected from six diverse robot platforms (i.e., Franka~\cite{franka_site}, UR5e~\cite{ur5e_site}, AgileX~\cite{AgileX_site}, ARX~\cite{arx2024lift}, Tien~Kung~\cite{tien_kung_site}, and Tian~Yi~\cite{xhumanoid2024tianyi}). 
Our manipulation tasks emphasize \nskills fundamental robotic skills and involve 1,139 distinct objects, with around 97K examples of common “pick-and-place” operations on different type of objects and over 50K instances of interactive manipulation tasks such as switching, dragging, and tool use (see Figure~\ref{fig:teaser}). 
In terms of scenario distribution, our dataset covers a variety of settings, including domestic scenes such as living rooms, bedrooms, kitchens, supermarkets, and children’s rooms, as well as industrial environments like logistics sorting facilities, biological laboratories, and industrial assembly lines, with roughly equal proportions of industrial and domestic scenarios.
\ours aggregates over 1K hours of operational experience, including 12K bimanual mobile manipulation trajectories enriched with tactile information.
Crucially, every trajectory is annotated with detailed natural language descriptions, enabling language-guided policy learning and supporting multimodal representation training for vision-language-action (VLA) models. Table~\ref{tab:dataset_cmp} presents a detailed comparison of RoboMIND 2.0 against a broad spectrum of contemporary datasets, including RT-1~\cite{brohan2022rt}, BC-Z~\cite{jang2022bc}, BridgeData V2~\cite{ebert2021bridge}, RoboSet~\cite{bharadhwaj2024roboagent}, BRMData~\cite{zhang2024empowering}, Dabb-E~\cite{shafiullah2023bringing}, Open X-Embodiment~\cite{o2024open}, AgiBot World~\cite{bu2025agibot}, and RoboCOIN~\cite{wu2025robocoin}, across key axes such as trajectory scale, task and skill diversity, embodiment configuration, multimodal sensing, detailed annotation, and sim-to-real alignment.

Moreover, like its predecessor RoboMIND 1.0~\cite{wu2024robomind}, \ours is collected in a unified and standardized experimental environment using a consistent teleoperation protocol and data recording pipeline, ensuring high data consistency and reproducibility. This standardized design effectively minimizes noise and bias introduced by environmental variations, providing a solid foundation for training robotic models with strong generalization capabilities across tasks and platforms.
More importantly, while RoboMIND 1.0~\cite{wu2024robomind} primarily focuses on single-arm manipulation in human-like domestic and industrial scenarios, \ours achieves a fundamental transition—from single-arm to bimanual coordinated operation. The current version systematically collects a large volume of bimanual manipulation trajectories in complex tasks, covering high-level skills that require two-handed coordination, such as grasping, assembly, switch operation, and object handover—behaviors that closely mirror natural human interaction patterns in the real world. Furthermore, our data collection extends beyond robotic systems equipped with simple parallel grippers; we also incorporate tasks performed using dexterous hands as end-effectors, enabling richer and more human-like manipulation behaviors.
This evolution not only enhances the dataset’s expressive power in terms of task complexity and interaction richness, but also provides a more challenging and practical foundation for research on bimanual coordination control, cross-morphology policy transfer, and embodied intelligence generalization. Combined with standardized data collection and fine-grained annotations, \ours offers a more comprehensive and normative resource for building robotic learning models that are generalizable, interpretable, and reproducible.

To further accelerate research and lower the barrier to entry, we emphasize the critical role of simulation in robotic development—particularly its ability to generate large-scale, high-quality data at a fraction of the cost of real-world collection. Physical data acquisition demands expensive hardware, extensive human supervision, and is prone to wear, damage, and safety constraints; in contrast, simulation enables rapid, safe, and repeatable data generation with near-zero marginal cost.
In alignment with this principle, we open-source all digital assets used in our data collection—including high-fidelity URDF models, scene layouts, and sensor configurations—and release a 20K-trajectory simulated dataset collected in simulation using two representative platforms: the Franka dual-arm gripper system and the Tien Kung dual-arm humanoid with dexterous hands. Critically, these simulated trajectories mirror the exact task structures, object configurations, and language instructions as the real-world data.

\ours comprises extensive long-horizon mobile manipulation trajectories—capturing temporally extended, semantically rich interactions across diverse environments. Given the scarcity of such data in existing benchmarks, a critical next step is to validate its effectiveness in enabling capable and generalizable policies.
Despite recent advances in VLA models, their performance on long-horizon mobile manipulation tasks remains limited, primarily due to the lack of large-scale, real-world datasets that capture temporally extended, semantically rich interactions across diverse environments~\cite{jiang2025galaxea,figure2024helix,shi2025hirobot,chen2025fast}. To address this gap and validate the utility of our newly collected long-horizon mobile manipulation data, we present \textbf{MIND-2}, a dual-process robotic manipulation framework that combines a slow, high-level planner (MIND-2-VLM) with a fast, low-level executor (MIND-2-VLA) to achieve robust long-horizon task completion in diverse real-world settings. MIND-2-VLM decomposes complex instructions into grounded, executable subtasks, while MIND-2-VLA executes these subtasks using egocentric vision, proprioception, and language guidance. Trained offline on large-scale real-world data via Implicit Q-Learning (IQL)~\cite{kostrikov2021offline}, MIND-2-VLA learns to imitate successful behaviors while avoiding failure modes by leveraging advantage-weighted regression. This integration of semantic planning and precise, reinforcement-learned control enables MIND-2 to generalize effectively across open-world manipulation tasks.

We conduct extensive experiments on both the construction of the large-scale \ours dataset and the slow/fast system MIND-2-VLM/VLA models. Regarding the \ours dataset, we perform detailed evaluations on data collected from robots with different configurations.
For assessment and benchmark analysis of our large-scale bimanual manipulation dataset \ours, we evaluate it with conventional imitation learning methods commonly used in single-robot manipulation, as well as VLA models that have shown strong generalization across diverse robotic tasks.
In our single-task imitation learning evaluation, we adopt two representative 2D-based approaches (ACT~\cite{zhao2023act} and UVA~\cite{li2025uva}) and two 3D-aware methods (DP3~\cite{ze20243d} and Dense Policy~\cite{su2025dense}). The results demonstrate that 3D imitation learning algorithms outperform 2D methods on tasks requiring bimanual coordination. This advantage likely stems from the richer spatial modeling in 3D, which enables more accurate representation of the visual dynamics involved in dual-arm interaction and synchronization.
In the context of multi-task VLA models, we conduct evaluations on the \ours dataset using $\pi_{0}$~\cite{black2024pi_0}, $\pi_{0.5}$~\cite{black2025pi_0.5}, HybridVLA~\cite{liu2025hybridvla}, and XR-1~\cite{fan2025xr}. The results demonstrate that XR-1~\cite{fan2025xr}, a cross-embodiment model, exhibits superior performance in bimanual manipulation tasks. Meanwhile, we also evaluate incorporating tactile signals as part of the robot's proprioceptive input to VLA models, which helps enable bimanual mobile manipulation tasks across diverse environments.
We then present a comprehensive analysis of the fast-slow MIND-2 system. Our evaluation focuses on long-horizon bimanual mobile manipulation tasks, including three collaborative scenarios involving the Tian Yi and AgileX robots across kitchen, supermarket, and industrial environments: tidying tableware, assisting at checkout, and performing material sorting. In these challenging settings, both standard imitation learning methods and existing VLA models underperform. The MIND-2 slow system (MIND-2-VLM) functions as a cloud-based “brain,” orchestrating high-level task stages and enabling coordinated control of heterogeneous robots. 
The MIND-2 fast system (MIND-2-VLA) is used for different robots to execute specific manipulation subtasks from MIND-2-VLM instruction.
Results demonstrate that the MIND-2 fast-slow system significantly outperforms single-task imitation learning and VLA baselines across all evaluated tasks. We evaluate object-level generalization in VLA models—arguably the most challenging aspect of real-world deployment—by training $\pi_{0.5}$~\cite{black2025pi_0.5} and XR-1~\cite{fan2025xr} on two UR5e dual-arm tasks and testing them with functionally equivalent but visually or geometrically novel objects.The results demonstrate that both $\pi_{0.5}$~\cite{black2025pi_0.5} and XR-1~\cite{fan2025xr} exhibit strong object-level generalization capabilities in bimanual manipulation tasks. The \ours dataset is publicly available on ModelScope at \url{https://modelscope.cn/datasets/X-Humanoid/RoboMIND2.0}.

In summary, we highlight several core contributions of RoboMIND 2.0:
\begin{itemize}
    \item \textbf{RoboMIND 2.0: A Large-Scale, Multimodal Bimanual Mobile Manipulation Dataset.}  
    We introduce \ours, a comprehensive robotic dataset comprising over 310K trajectories across 759 tasks and 129 skills, collected from six diverse dual-arm platforms (including mobile manipulators and humanoids) in both domestic and industrial environments. \ours is the first open dataset to jointly support bimanual coordination, mobile manipulation, dexterous hands, and high-fidelity tactile sensing.

    \item \textbf{Multi-Dimensional Diversity Beyond Existing Benchmarks.}  
    Unlike prior datasets that focus on only a single aspect of diversity (e.g., objects, tasks, or embodiments), \ours provides simultaneous coverage across robot morphology, environment, task semantics, failure modes, and multimodal sensing, including synchronized vision, proprioception, force-torque, and tactile feedback, enabling truly generalizable policy learning.

    \item \textbf{High-Fidelity Digital Twin for Sim-to-Real Transfer.}  
    We release a photorealistic simulation environment with exact replicas of real-world assets and a 20K-trajectory simulated dataset aligned with real data in task structure, language instructions, and object configurations. This enables cost-effective, scalable training and validates simulation as a critical engine for embodied AI.

    \item \textbf{MIND-2: A Dual-Process Framework for Long-Horizon Bimanual Tasks.}  
    To leverage our dataset, we propose MIND-2, a slow-fast robotic architecture combining a high-level vision-language planner (MIND-2-VLM) and a low-level vision-language-action executor (MIND-2-VLA). MIND-2 achieves robust long-horizon bimanual mobile manipulation in complex real-world scenarios where existing VLA models fail.
    \item \textbf{Empirical Validation Across Imitation Learning, VLA Models, and Cross-Embodiment Generalization.} 
    Through extensive experiments, we demonstrate that 3D-aware imitation learning methods significantly outperform 2D counterparts in bimanual manipulation tasks, owing to their enhanced spatial reasoning capabilities. We further show that modern vision-language-action (VLA) models, such as XR-1 and $\pi_{0.5}$, exhibit strong cross-embodiment and object-level generalization, successfully transferring skills across diverse robot morphologies and handling functionally equivalent but visually or geometrically novel objects. Moreover, incorporating tactile signals into the policy input consistently yields substantial performance gains, underscoring the critical role of physical interaction feedback in dexterous manipulation. Finally, our proposed MIND-2 framework proves highly effective in challenging, long-horizon scenarios involving collaborative, multi-robot coordination across domestic and industrial environments, significantly outperforming standard imitation learning and existing VLA baselines.
\end{itemize}

\section{Related Work}

\subsection{Policy Learning for Robotic Manipulation}

The paradigm of robotic manipulation has shifted significantly from specialized control relying on states to generalizable learning grounded in vision. 
Early approaches primarily relied on Reinforcement Learning (RL) based on states to solve specific control tasks~\citep{schulman2017proximal,haarnoja2018soft,fujimoto2018addressing,andrychowicz2020learning,joshi2020robotic,yarats2021mastering}. 
However, the necessity for robots to operate in unstructured environments spurred the integration of high-dimensional visual observations~\citep{mo2021where2act,eisner2022flowbot3d,fang2023anygrasp}.
Recent advancements have demonstrated that Imitation Learning (IL)~\citep{deng2018learning,zhao2023learning,chi2023diffusion_policy,bharadhwaj2024roboagent,wu2024swbt,buamanee2024bi,zare2024survey,fu2024mobile}, which allows robots to acquire diverse skills by mimicking expert demonstrations, offers a stable and scalable learning paradigm. 
Drawing inspiration from the success of image synthesis models~\citep{ho2020ddpm,rombach2022stable_diffusion,esser2024scaling,labs2025flux1kontextflowmatching}, generative models have further enhanced the representation capability of robotic manipulation policies. 
Diffusion policies~\citep{chi2023diffusion_policy,pearce2023imitating_diffusion,reuss2023goal,wu2024discrete,ryu2024diffusion,li2024crossway} treat policy learning as a conditional generation problem by transforming Gaussian noise into coherent action sequences. 
These approaches are particularly effective at fitting the multimodal action patterns inherent in human demonstrations. 
Recent methods have been further extended to 3D workspaces by utilizing point clouds and multi-view representations to enhance spatial precision~\citep{ze20243d,ke20243d,gervet2023act3d,goyal2023rvt,shridhar2023perceiver,jia2024lift3d,zhang2025flowpolicy,zhu2024spa,ze2024humanoid_manipulation}. 
Despite these developments, most current methods are primarily applied within simulation or real-world experimental settings, limited to single or a few tasks. 
Consequently, the range of executable tasks remains narrow while both generalization and robustness are often insufficient. 
There remains a notable lack of unified and large-scale open source robotic manipulation datasets to provide the general foundation knowledge necessary for enhancing model learning capabilities.
As a large-scale open source dataset comprising six distinct robot embodiments and 310K dual-arm manipulation trajectories, \ours~ 
can significantly enhance the foundational general capabilities of robotic policies by providing the diverse data necessary for robust learning.

\begin{table*}[t]
  \centering
  \setcounter{footnote}{1}
  \caption{
    Comparison to existing real-world datasets for robot manipulation. 
    We report the number of unique multi-view trajectories and highlight the advantages of \ours in \orange{orange}. 
    \small{
      \dgray{$^\ddag$non-robot, tool-based data collections.}
      \gray{$^\S$not a dataset in itself, but an \emph{aggregation} of existing datasets.}
    }
  }
  \resizebox{1.0\textwidth}{!}{
  \begin{tabular}{l c c c c c c c c c l}
    \toprule
    \textbf{Dataset} & \textbf{Trajectory} & \textbf{Task}  & \textbf{Skill}  & \textbf{Dexterous Hand}& \textbf{Detailed Annotation}  &  \textbf{Mobile Manipulation}  & \textbf{Embodiment} & \textbf{Tactile Information} & \textbf{Digital Twin} & \textbf{Collection} \\
    \midrule
    RT-1~\cite{brohan2022rt}                  & 130k  & 700 & 8    & \xmark & \xmark & \xmark & 1 & \xmark & \xmark & Human Teleoperation \\
    BC-Z ~\cite{jang2022bc}                   & 26k   & 100 & 3    & \xmark & \xmark & \xmark & 1 & \xmark & \xmark & Human Teleoperation \\
    BridgeData~V2~\cite{walke2023bridgedata}  & 60.1k & n/a & 13   & \xmark & \xmark & \xmark & 1 & \xmark & \xmark & 85\% Human / 15\% Scripted \\
    RoboSet~\cite{bharadhwaj2024roboagent}   & 98.5k & 38  & 6    & \xmark & \xmark & \xmark & 1 & \xmark & \xmark & 30\% Human / 70\% Scripted \\
    RH20T~\cite{fang2024rh20t}                & 13k   & 140 & 33   & \xmark & \xmark & \xmark & 1 & \xmark & \xmark & Human Teleoperation \\
    DROID~\cite{khazatsky2024droid}            & 76k   & n/a & 86   & \xmark & \xmark & \xmark & 1 & \xmark & \xmark & Human Teleoperation \\
    BRMData~\cite{zhang2024empowering}         & 0.5k  & 10  & 7     & \xmark & \xmark & \xmark & 1 & \xmark & \xmark & Human Teleoperation \\
    \dgray{Dobb-E$^\ddag$}~\cite{shafiullah2023bringing}    & \dgray{5.6k} & \dgray{109} & \dgray{6}        & \dgray{\xmark}  & \dgray{\xmark}  & \dgray{\xmark}  & \dgray{1} & \dgray{\xmark}  & \dgray{\xmark}  & \dgray{Human Tool-based}   \\
    \gray{Open~X-Embodiment$^\S$}~\cite{o2024open}       & \gray{1.4M}  & \gray{160k} & \gray{217}    & \gray{\xmark}   & \gray{\xmark}   & \gray{\xmark}   & \gray{22}  & \gray{\xmark}   & \gray{\xmark}   & \gray{Dataset Aggregation} \\
    RoboMIND~\cite{wu2024robomind}
    & 107K      & 479     & 38    & \cmark & \cmark & \xmark & 4 & \xmark & \cmark & Human Teleoperation  \\
    AgiBot World~\cite{bu2025agibot} & 1M & 217 & 87  & \cmark & \cmark & \xmark & 1 & \xmark & \xmark & Human Teleoperation \\
    Open Galaxea~\cite{jiang2025galaxea} & 50k & 150 & 58  & \xmark & \cmark &\cmark & 1 & \xmark & \xmark & Human Teleoperation \\
    RoboCOIN~\cite{wu2025robocoin} & 180k & 421 & 36  & \cmark & \cmark & \cmark & 15 & \xmark & \xmark & Human Teleoperation \\
     \cmidrule(lr){1-11}
       \ours
    & \orange{\ntrajs}      & \orange{\ntasks}     & \orange{\nskills}    & \orange{\cmark} & \orange{\cmark} & \orange{\cmark} & 6         & \orange{\cmark} & \orange{\cmark} & Human Teleoperation        \\
    \bottomrule
  \end{tabular}}
  \setcounter{footnote}{0}
  \captionsetup{width=\textwidth}
  \label{tab:dataset_cmp}
\end{table*}

\subsection{Large-Scale Robot Manipulation Datasets}

The development of generalist robotic policies capable of mastering manipulation tasks in unstructured environments depends fundamentally on the availability of high-quality training data. 
Although general-purpose simulators~\citep{coumans2016pybullet,gazebo,isaacgym,isaacsim,SAPIEN} and photorealistic simulation environments~\citep{AI2-THOR,chang2017matterport3d,gu2023maniskill2,habitat,tao2024maniskill3} offer scalable platforms for policy learning, the gap between simulation and reality frequently compromises the transferability of these policies to hardware and task scenarios in the real world. 
Consequently, research has pivoted toward acquiring data directly from the real world because such data originates from actual application scenarios and represents the most valuable and premium resources. 
Early efforts utilized automated scripts based on rules or expert policies specific to a domain~\citep{pinto2016supersizing,gupta2018robot,levine2018learning,cabi2019scaling,dasari2020robonet,kalashnikov2021mt}. 
Human teleoperation has since emerged as a universal method for capturing complex, dexterous, and extended behaviors~\citep{mandlekar2018roboturk,sharma2018multiple,ebert2021bridge,brohan2022rt,jang2022bc,walke2023bridgedata,bharadhwaj2024roboagent,fang2024rh20t}.

Table~\ref{tab:dataset_cmp} presents the comparison and differences between \ours and other representative public robotic manipulation datasets collected in the real world. 
While these datasets possess high quality, they typically exhibit prominence only in individual dimensions and lack overall versatility or massive scale across multiple metrics. 
For instance, BridgeData V2~\citep{walke2023bridgedata} and RoboSet~\citep{bharadhwaj2024roboagent} provided critical trajectory counts exceeding 50K but remained limited to a narrow set of skills, containing only 13 and 6 skills, respectively. 
In contrast, datasets featuring diverse skills like RH20T~\citep{fang2024rh20t} contain 33 skills but only 13K trajectories, which lack the data volume required for pretraining large foundation models. 
DROID~\citep{khazatsky2024droid} contains a higher count of 86 skills and collects 76K demonstration trajectories via human teleoperation, yet it only comprises data for the single-arm Franka robot.
Other works have focused on massive data aggregation and the unified alignment of data storage. 
Open X-Embodiment~\citep{o2023open} has made a significant effort to unify existing robot datasets into a standardized format by incorporating data from diverse robots collected through collaboration among 21 institutions. 
Following this, ARIO~\citep{wang2024all} further integrates data from the real world and simulation into a standard format to bridge the gaps in existing data resources. 
Building on this foundation, the most recent works have pushed the boundaries of dataset magnitude and quality even further. 
AgiBot World~\citep{bu2025agibot} and Galaxea Open World~\citep{jiang2025galaxea} have achieved immense data volume on specific robotic embodiments to master diverse tasks, scenes, and skills. 
At the same time, RoboMIND~\citep{wu2024robomind} and RoboCoin~\citep{wu2025robocoin} also enhanced embodiment breadth by covering heterogeneous embodiments, containing 4 and 15 robot types, respectively.
However, a significant challenge remains as current datasets tend to focus either on deep mastery of a single embodiment, which limits their generalization to new embodiments, or on broad coverage of multiple embodiments that dilutes the density of tasks per robot. 
Furthermore, a critical limitation across nearly all existing massive benchmarks is the reliance on observations that are purely visual, which neglects the tactile feedback that is indispensable for robust manipulation involving rich contacts. 
Bridging these gaps requires a dataset that simultaneously ensures embodiment diversity, task richness, and multimodal data collection. 
To this end, we introduce \ours, a multimodal, bimanual mobile manipulation dataset for generalizable embodied intelligence. 
Unlike predecessors that trade off between embodiment variety and task depth, \ours~ comprises {\ntrajs} expert trajectories across {\ntasks} tasks and {\nobjs} object classes. 
By explicitly integrating diverse embodiment configurations with tactile sensory data, it establishes a comprehensive and high-quality standard designed to advance next-generation dexterous policy learning across multiple embodiments.

\subsection{Large-Scale Vision-Language-Action Models}

Learning robotic policies from large-scale data has gradually become a cornerstone of modern embodied intelligence. 
One prominent class of methods utilizes human manipulation data sourced from the internet~\citep{goyal2017something,damen2018scaling,damen2022rescaling,grauman2022ego4d}. 
These approaches focus on transferring knowledge from human manipulation to robotic actuation, encompassing the learning of robust human-robot representations~\citep{nair2022r3m,bahl2023affordances}, the extraction of embodiment-agnostic priors~\citep{mandikal2022dexvip,kannan2023deft}, and the control of dexterous hands~\citep{mandikal2021learning,wu2023learning}. 
Recent frameworks such as VPP~\citep{hu2024video}, LAPA~\citep{ye2024latent}, and GR-2~\citep{cheang2024gr} further exploit large-scale actionless video pre-training to capture task semantics, aiming to align visual dynamics with robotic action spaces. 
However, effectively transferring this human-centered data to precise low-level robotic control remains a significant challenge, necessitating the integration of high-quality robotic data.

In response to this challenge, there has been a surge in work developing real-world robotic manipulation datasets. 
Early efforts like BridgeData~\citep{ebert2021bridge,walke2023bridgedata} focused on controlled laboratory environments, while recent initiatives such as Open X-Embodiment~\citep{o2024open}, DROID~\citep{khazatsky2024droid}, RoboMIND~\citep{wu2024robomind}, and AgiBot World~\citep{bu2025agibot} consolidate trajectories across diverse robot hardware and scenes. 
Researchers are increasingly investigating how to aggregate these large-scale and heterogeneous datasets to elevate the foundational capabilities of models, shifting the paradigm from specialized imitation learning on narrow demonstrations to the training of generalist foundation models. 
Concurrently, the emergence of Vision-Language-Action (VLA) models has established unified pre-training paradigms that integrate vision, language, and action~\citep{zitkovich2023rt,li2023vision,li2023manipllm,liu2024robomamba,kim2024openvla}. 
By fine-tuning Vision-Language Models (VLMs) combined with an action head module on robotic trajectory data, VLA models enable agents to interpret natural language commands~\citep{mao2016generation,liu2023visual,2023_url_ShareGPT,awadalla2023openflamingo} and predict low-level robotic control signals directly~\citep{brohan2022rt,mees2022calvin,sermanet2023robovqa,wen2024tinyvla,driess2023palm}. 
A prominent approach involves co-training VLMs on both internet-scale corpora and robotic trajectories, as demonstrated by RT-2~\citep{zitkovich2023rt}, PaLM-E~\citep{driess23palme}, and RoboFlamingo~\citep{li2023vision}, which transfer semantic reasoning to physical control.

Leveraging massive robotic datasets, numerous recent studies focus on augmenting VLA models with richer physical knowledge and diverse capabilities~\citep{zhang2023crossformer,kim2024openvla,wang2024scaling,black2024pi_0,liu2024rdt,wen2025tinyvla,bjorck2025gr00t,liu2025hybridvla,li2025switchvla,fan2025diffusion,liu2025mla,zheng2025x,bi2025h}. 
Architectures such as CrossFormer~\citep{zhang2023crossformer}, Octo~\citep{team2024octo}, HPT~\citep{wang2024scaling}, and $\pi_0$~\citep{black2024pi_0} construct embodiment-general representations that achieve robust cross-embodiment transfer, which allows models to control distinct robots effectively. 
Building on these generalist foundations, subsequent works have introduced specialized mechanisms to further refine specific capabilities.
For instance, $\pi_{0.5}$~\citep{intelligence2025pi_05} and FSD~\citep{yuan2025seeing} integrate large-scale image-text pre-training to improve semantic understanding and visual grounding. 
Diffusion-VLA~\citep{wen2025diffusionvla} utilizes diffusion processes for multimodal action generation, while InstructVLA~\citep{yang2025instructvla} focuses on improving fidelity to natural language instructions. 
Furthermore, CoT-VLA~\citep{zhao2025cot} and SpatialVLA~\citep{qu2025spatialvla} integrate Chain-of-Thought reasoning and spatial awareness,s respectively. 
Approaches like XR-1~\citep{fan2025xr} and VPDD~\citep{he2024learning} introduce discrete latent codes and diffusion modeling to explicitly align human dynamics from video data with robotic motion derived from large-scale robotic data.
Building upon these advancements, we propose the fast-slow MIND-2 system. 
In this framework, the slow system (MIND-2-VLM) serves as the brain for high-level task planning and cooperative control of heterogeneous robots, while the fast system (MIND-2-VLA) executes fine-grained control for specific tasks. 
Crucially, the instruction-following capabilities, multi-task performance, and multi-scene generalization of these models are not merely products of architectural design but are fundamentally driven by the scale and diversity of pre-training datasets. 
RoboMIND 2.0, as a large-scale cross-embodiment robotic manipulation dataset, significantly enhances the foundational manipulation capabilities of various VLA models. 
Moreover, it includes large-scale tactile data, thereby opening new avenues for research into tactile-fused training paradigms.

\section{ \ours Dataset Construction}

\ours dataset is a high-quality dataset encompassing  \textbf{\nskills} robotic manipulation skills, covering diverse scenarios in both home living environments and industrial settings.
Our dataset contains a total of \textbf{\ntrajs} dual-arm manipulation trajectories from various robot configurations. These configurations include dual-arm setups with human-like arrangements using Franka~\cite{franka_site} and UR5~\cite{ur5e_site} robots, wheeled mobile platforms such as Tian~Yi~\cite{xhumanoid2024tianyi}, AgileX~\cite{AgileX_site}, and ARX~\cite{arx2024lift}, as well as the humanoid robot Tien Kung~\cite{tien_kung_site}. Among the collected data, \textbf{290K} trajectories involve dual-arm collaborative operations with the arms fixed on a tabletop, while \textbf{20K} trajectories consist of mobile dual-arm operations in both industrial and home environments. The dataset covers tasks in common household scenarios such as kitchens, bedrooms, living rooms, and bookshelves, as well as industrial applications including logistics sorting, hazardous inspection, and physical and chemical experiments. These tasks involve over \textbf{\nobjs} distinct operational objects and span \textbf{\nskills} common robotic manipulation skills. 
Additionally, we have open-sourced the simulation assets for all the included datasets. 
In the following, we will introduce the \ours dataset in detail, covering how data was collected and quality-checked for different robot configurations and how the  \ours dataset
is annotated.
  
  
  
  

\subsection{Data Collection}

\ours is a dataset encompassing various robot configurations, including humanoid-style dual-arm robots such as the Franka~\cite{franka_site} and UR5e~\cite{ur5e_site},  dual-arm mobile robots like the ARX~\cite{arx2024lift} and AgileX~\cite{AgileX_site}, as well as the humanoid robots including Tien~Kung~\cite{tien_kung_site} and Tian~Yi~\cite{xhumanoid2024tianyi}.
Different types of robots  have specific teleoperation devices for collecting robot data. In the following section, we provide a detailed explanation of the teleoperated data collection process for different types of dual-arm robots.

\begin{figure}[h]
    \centering
    \includegraphics[width=0.98\textwidth]{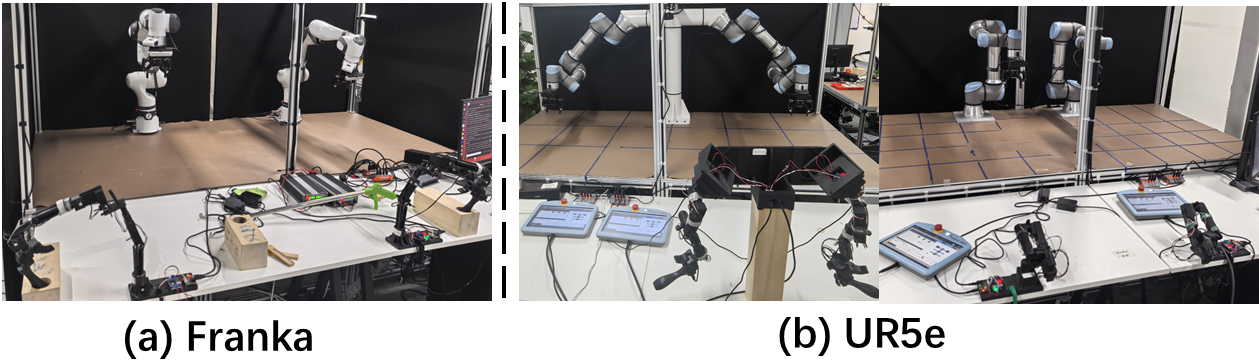}
    \caption{\textbf{Collection platforms of Franka and UR5e.} Collect a robotic manipulation dataset by controlling the dual-arm system (Franka and UR5e) via HACTS.}
    \label{fig:htact}
\end{figure}

\textbf{Franka and UR5e.}
As shown in the Figure~\ref{fig:htact}, we place two single-arm Franka robots side by side on a workbench to collect data for bimanual collaborative tasks. The dual UR5e arms are deployed in two configurations: one with both arms mounted parallel on the workbench, and the other arranged in a humanoid-like bimanual layout, enhancing morphological diversity.
Data collection is performed using the HACTS~\cite{xu2025hacts} teleoperation system, which is a low-cost, lightweight platform that supports kinematically equivalent adaptation and bidirectional synchronized control.
The hardware uses Dynamixel servos (XL430-W250-T for high-load joints and XL330-M288-T for the grippers and others), a 3D-printed PLA frame, a 12V/5A power supply, and a WaveShare servo adapter and supports kinematically equivalent adaptation. 
The software relies on the DYNAMIXEL API for motor control, enables bidirectional robot–hardware synchronization via offset calibration, and allows switching between autonomous and human-guided modes using a foot pedal.
HACTS enable motion control by mapping the movements of the slave arms to the master arms. During data collection, operators perform the dual-arm robotic manipulation tasks by controlling two independent slave arms. 

\begin{figure}[h]
    \centering
    \includegraphics[width=0.98\textwidth]{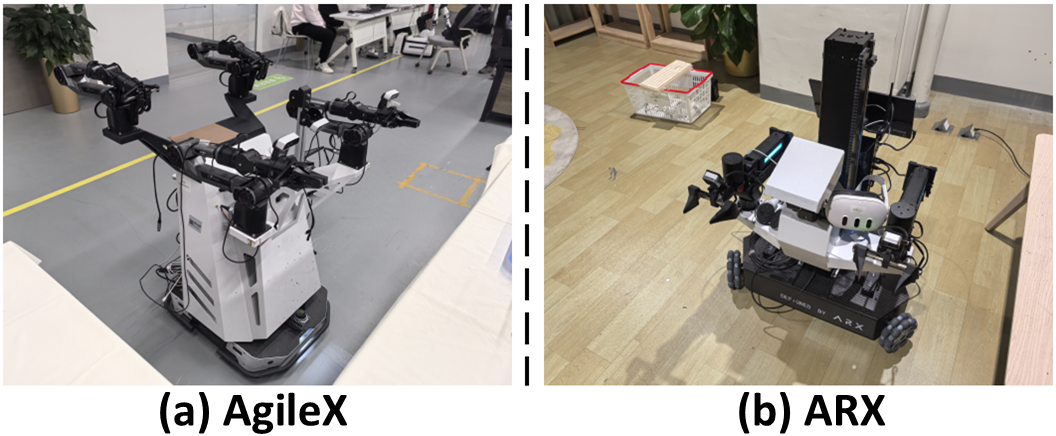}
    \caption{\textbf{Collection platforms of AgileX and ARX.} We use a VR headset to control the ARX robot for data collection, and employ a slave arm to teleoperate the master arm for gathering the AgileX manipulation dataset. }
    \label{fig:agliex}
\end{figure}

\textbf{AgileX and ARX.}
For the AgileX, we employ a bilateral teleoperation system similar to the Mobile ALOHA~\cite{zhao2023learning}, directly collecting the dataset on the robot platform to ensure the naturalness and precision of the demonstrated actions.
Figure~\ref{fig:agliex} shows that for the Agilex arm data collection, we employ a teleoperation structure using an auxiliary robotic arm to control the main robotic arm.
To collect mobile platform data for the AgileX robot, we captured real-time linear and angular velocity measurements by physically pushing the robot's base during operation.
For the ARX arm data, we apply VR headsets to capture the motion of human arms, and then mapped the recorded movement data to the dual-arm robot for teleoperated manipulation.
For the ARX mobile manipulation data, we used a VR controller to command the motion direction and speed of the mobile base. 
The base then moves accordingly, and its real-time movement trajectories, including various directions and velocities, are recorded.
Figure~\ref{fig:agliex} shows that  the data collection operator is wearing a VR headset to teleoperate the ARX dual-arm robot and collect the dataset for bimanual robotic tasks.

\begin{figure}[h]
    \centering
    \includegraphics[width=0.98\textwidth]{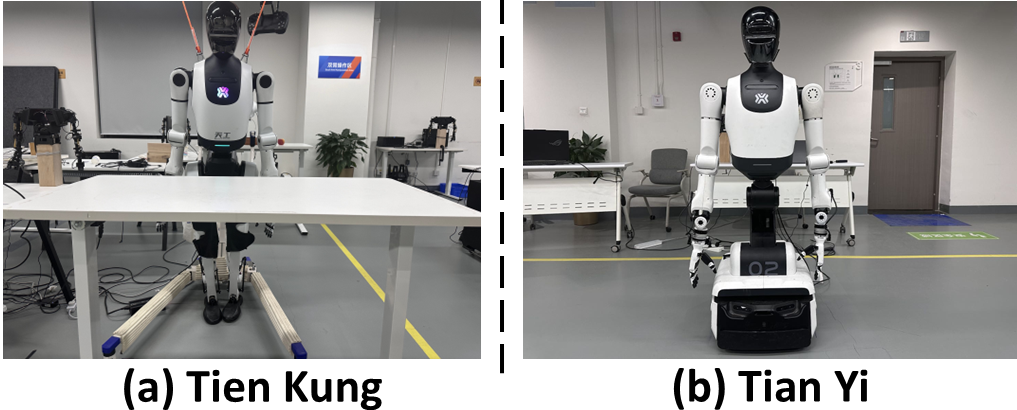}
    \caption{Visualization of Tian Yi and Tien Kung robots.}
    \label{fig:tianyi}
\end{figure}

\textbf{Tien Kung and Tian Yi.}
As shown in the Figure~\ref{fig:tianyi}, the Tien Kung robot is a humanoid dual-arm robot. The Tian Yi robot features a humanoid upper body with dual arms, while its lower body is equipped with a wheeled base for mobility.
For the humanoid robot Tien Kung,  data collectors  wear motion capture suits to record joint movements, which are then mapped to the robot to enable robotic manipulation. For the dual-arm mobile robot with a wheeled base Tian Yi, we collect datasets using two human operators. One operator control the dual-arm manipulation using a HACTS device, while the other collect mobility data by physically pushing the robot to record the base's angular and linear velocities, thereby capturing the robot's motion dynamics.

To optimize storage efficiency and facilitate dataset organization, we consolidate each collected trajectory, encompassing multi-view RGB-D data, robot proprioceptive state information, specific end-effector state information, and teleoperation body state information, into a single HDF5 format file.
For the AgileX mobile manipulation dataset from \ours, we also provide tactile information from both the left and right robot arms in the HDF5  files.

\subsection{Data Inspection}

\begin{figure}[h]
  \centering
  \includegraphics[width=0.98\columnwidth]{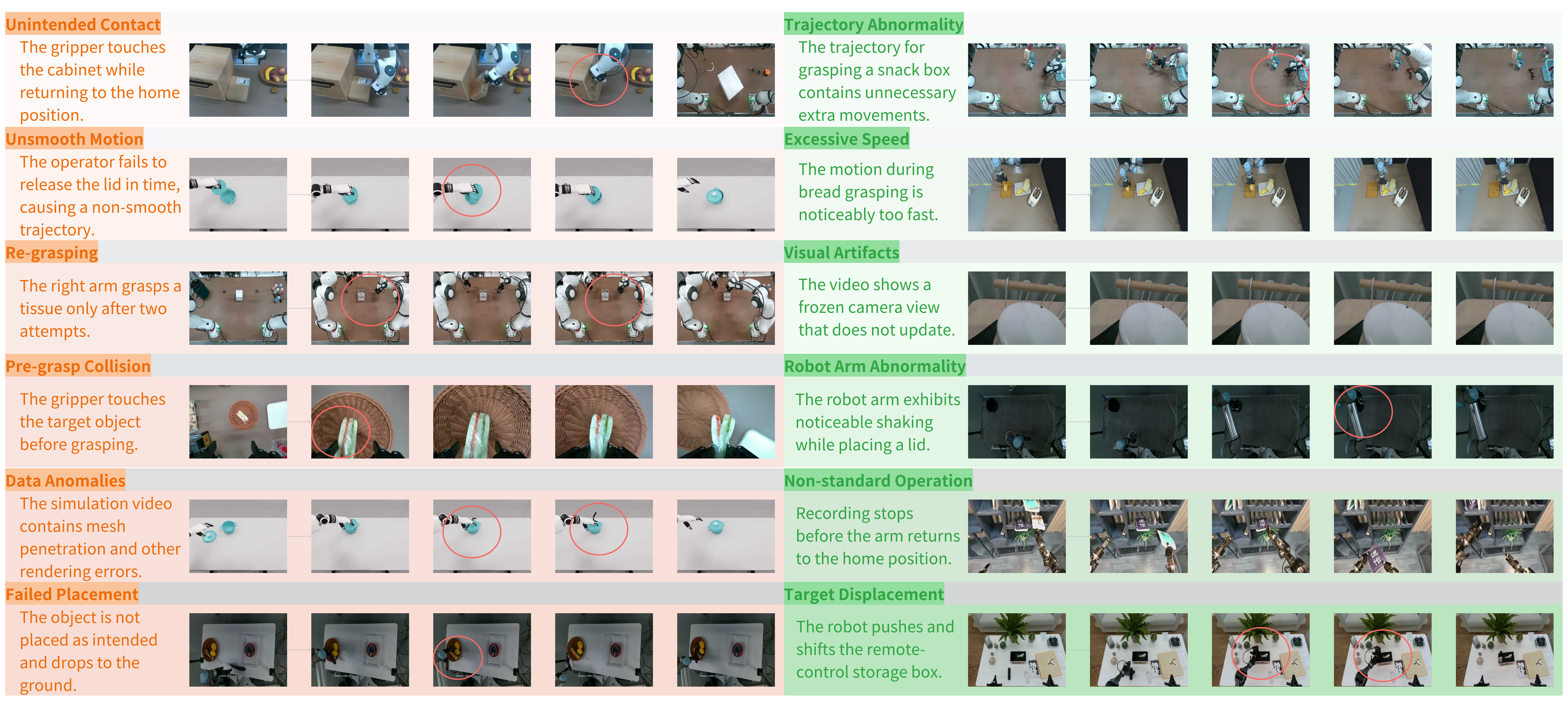}
  \caption{\textbf{An example of data inspection categories.} Illustration of the twelve data inspection categories used in our quality-control workflow, including unintended contact, motion irregularities, sensing artifacts, and task-level execution failures.}
  \label{fig:inspection}
\end{figure}

All data in our dataset is collected through real-time teleoperation. Because operators must continuously control dual robots over extended sessions, data quality can be affected by natural human factors such as fatigue, inconsistent habits, momentary distraction, or external disturbances. To mitigate these issues, we employ a rotating-shift schedule and provide a comfortable, low-noise working environment to help operators maintain stable and reliable performance.

Building on this, we implement a centralized multi-stage data–inspection workflow designed to ensure consistency and high fidelity across all trajectories. At the end of each collection day, raw logs and videos are automatically organized into a structured daily report. A random subset is first sampled for preliminary checking; if issues are detected, they are compiled into a ``daily issue report'' and immediately fed back to the collection team for timely adjustment. The remaining trajectories are assigned to dedicated quality-control staff, who review each video and annotate all irregularities. Once all issues are processed by discarding problematic sequences or re-collecting missing data, the corresponding sessions are marked as verified and released for downstream use.

During detailed inspection, each trajectory is evaluated against a set of twelve quality criteria, each capturing a common failure mode in teleoperated demonstrations. These criteria correspond to the examples shown in Figure~\ref{fig:inspection} and are formally defined as follows:

\vspace{4pt}
\begin{itemize}
  \item \textbf{Unintended Contact.}
    \begin{itemize}
      \item \emph{Definition:} Unintentional physical contact with non-target objects that introduces disturbances or contaminates the recorded trajectory.
      \item \emph{Manifestation:} The gripper touches surrounding structures, such as cabinets or containers, during routine motion or while returning to the home position.
    \end{itemize}

  \item \textbf{Unsmooth Motion.}
    \begin{itemize}
      \item \emph{Definition:} Discontinuous or jerky motions caused by operator hesitation or lack of coordination, leading to irregular trajectories or duplicated segments.
      \item \emph{Manifestation:} Delayed release, repeated micro-adjustments, or abrupt transitions between motion phases.
    \end{itemize}

  \item \textbf{Re-grasping.}
    \begin{itemize}
      \item \emph{Definition:} Failure to grasp the object on the first attempt, requiring multiple attempts and reducing temporal consistency.
      \item \emph{Manifestation:} Two or more grasp attempts before successfully securing the object.
    \end{itemize}

  \item \textbf{Pre-grasp Collision.}
    \begin{itemize}
      \item \emph{Definition:} Unnecessary collision with the target object or nearby obstacles before the intended grasp, disturbing the scene state.
      \item \emph{Manifestation:} The gripper bumps into the object or neighboring items prior to initiating the grasp.
    \end{itemize}

  \item \textbf{Data Anomalies.}
    \begin{itemize}
      \item \emph{Definition:} Sensing or rendering corruptions, including frame drops, freezing, or simulation artifacts that impair data usability.
      \item \emph{Manifestation:} Stuttering playback, frozen frames, or mesh penetration in simulated environments.
    \end{itemize}

  \item \textbf{Failed Placement.}
    \begin{itemize}
      \item \emph{Definition:} The object is not placed stably or accurately at the designated location due to misalignment or improper release.
      \item \emph{Manifestation:} The object falls, tilts, or lands outside the intended placement region.
    \end{itemize}

  \item \textbf{Trajectory Abnormality.}
    \begin{itemize}
      \item \emph{Definition:} Significant deviation from smooth, expected trajectories, including unnecessary detours or path irregularities.
      \item \emph{Manifestation:} Extra loops, exaggerated corrections, or irregular end-effector paths not aligned with task intent.
    \end{itemize}

  \item \textbf{Excessive Speed.}
    \begin{itemize}
      \item \emph{Definition:} Motions executed at unrealistically high speeds, degrading control fidelity and physical realism.
      \item \emph{Manifestation:} Rapid grasping or placement actions far exceeding typical human teleoperation speeds.
    \end{itemize}

  \item \textbf{Visual Artifacts.}
    \begin{itemize}
      \item \emph{Definition:} Camera-level distortions or failures, such as flickering, color shifts, or frozen viewpoints.
      \item \emph{Manifestation:} Persistent color anomalies or static views that do not update.
    \end{itemize}

  \item \textbf{Robot Arm Abnormality.}
    \begin{itemize}
      \item \emph{Definition:} Mechanical or control irregularities that result in unstable behavior, including shaking or oscillation.
      \item \emph{Manifestation:} Visible vibration or instability during manipulation or placement.
    \end{itemize}

  \item \textbf{Non-standard Operation.}
    \begin{itemize}
      \item \emph{Definition:} Violations of the demonstration protocol, particularly failure to return to the home pose followed by a brief pause before termination.
      \item \emph{Manifestation:} Recording stops before the home pose is reached, or the pause duration is insufficient or excessively long.
    \end{itemize}

  \item \textbf{Target Displacement.}
    \begin{itemize}
      \item \emph{Definition:} Unintended movement of objects expected to remain static, caused by incidental collisions.
      \item \emph{Manifestation:} The robot inadvertently pushes or shifts items such as bins, boxes, or containers.
    \end{itemize}
\end{itemize}

For each violation, inspectors record precise timestamps and short textual descriptions. These annotations guide data filtering and help identify systematic issues, such as recurring pre-grasp collisions that indicate poor gripper alignment or repeated excessive-speed violations that suggest operator overcompensation. To operationalize these criteria at the trajectory level and make the procedure reproducible, we formalize the inspection protocol as a three-stage workflow:

\begin{itemize}
  \item \textbf{Initial Inspection:} Quickly scan videos to detect major technical issues such as frame loss or frozen views.
  \item \textbf{Detailed Inspection:} Review videos frame-by-frame or in slow motion to identify any of the twelve failure modes described above.
  \item \textbf{Data Filtering and Issue Logging:} Document timestamps and descriptions for all non-compliant data and categorize each sequence for discarding, correction, or re-collection.
\end{itemize}


\subsection{Data Annotation}

Although visual and robot proprioceptive information can be extracted directly from the collected videos and trajectories, we need to provide better semantic information from the data to aid model training.
For each collected mobile manipulation task, we provide detailed natural language annotations for every stage of the task.
In our fast-slow system, we first segment the long-horizon mobile manipulation task datasets into phase-specific sub-tasks based on stage-level language descriptions. These segmented datasets, along with their corresponding language instructions, are used to train the fast system VLA model. The slow VLM system then monitors task progress and determines when to switch between stages by generating or triggering the next high-level language command. This hierarchical cooperation enables the robot to successfully execute complex, long-horizon manipulation tasks.

In this work, we  annotate each successful robot motion trajectory, which is contained in long-horizon manipulation tasks. In the segmentation of long-horizon robotic tasks into sub-tasks, this work adopts a semantic action-based partitioning criterion: the transition between navigation and manipulation serves as the boundary for sub-action segmentation.
Specifically, navigation actions are defined as \textit{``Go to [location]''} or \textit{``Stop in front of [object]''}, while all manipulation actions occurring between two consecutive navigation actions are grouped into a single, coherent operation sequence and merged into one sub-task unit.
This segmentation strategy follows the spatiotemporal coherence and semantic integrity of task execution, ensuring that each sub-task has a clear functional objective. All annotations are generated automatically using the large language model Gemini 2.5 Pro~\cite{comanici2025gemini}, leveraging its strong contextual understanding and reasoning capabilities to perform semantic parsing and structured annotation of raw action sequences, thereby enabling efficient, consistent, and semantically accurate sub-task decomposition. We then use these annotated results as exemplars to guide Gemini in performing annotations based on this reference pattern.

\begin{wrapfigure}{r}{0.6\textwidth}
  \centering
  \vspace{-2em}
  \includegraphics[width=0.6\textwidth]{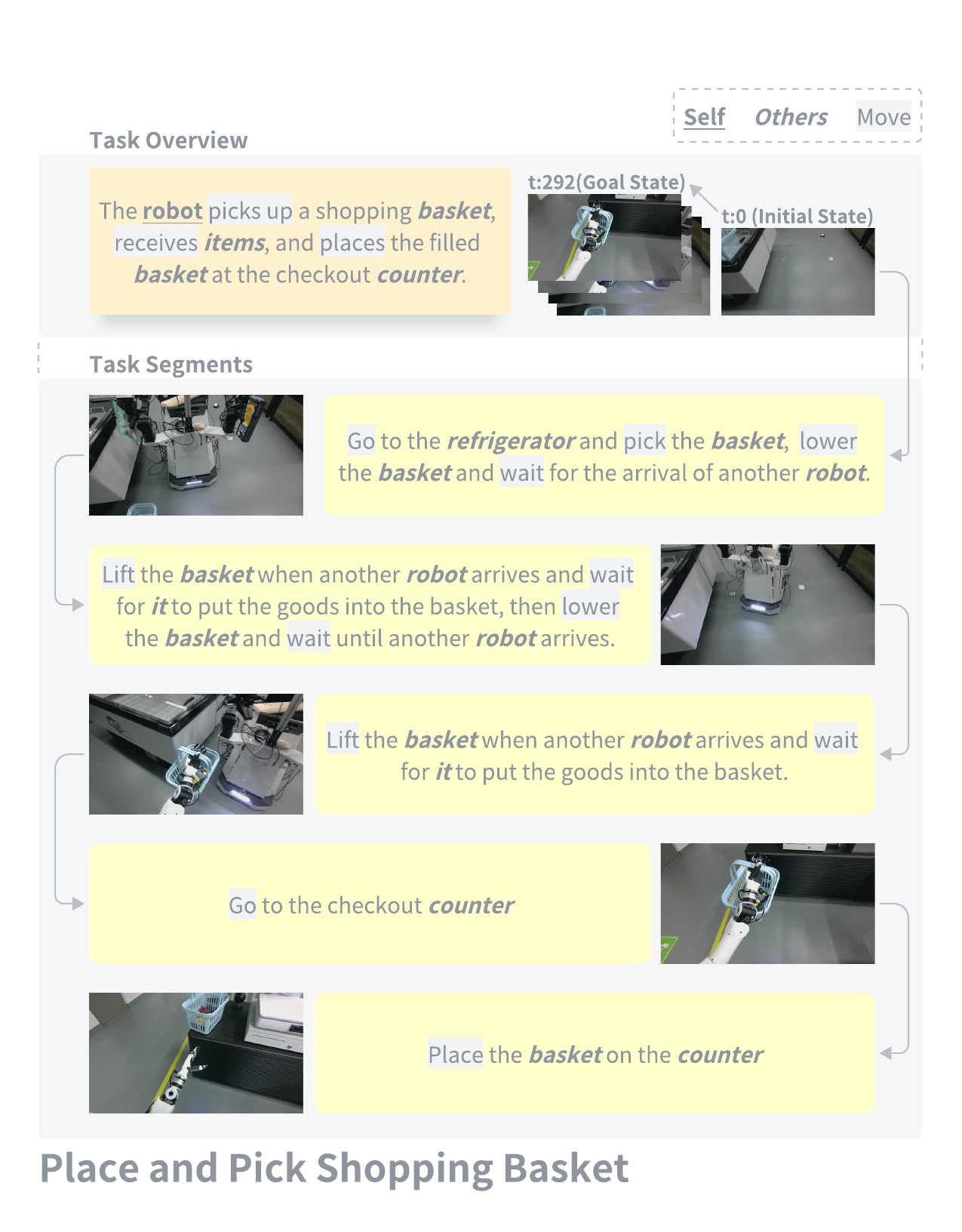}
  \caption{
    \textbf{Examples of language description annotation.}
    The video of the robotic arm placing the apple in the drawer is divided into six segments using Gemini.
    The language descriptions provided for each segment were initially generated by Gemini and subsequently manually refined.
  }
  \label{fig:ln} 
  \vspace{-2em}
\end{wrapfigure}

This thorough process enhances the precision and reliability of the language annotations for the collected trajectories. 
Figure~\ref{fig:ln} shows the language annotation for Tian Yi’s task in a dual-robot collaboration scenario. The task involves an AgileX dual-arm robot picking up bananas and chili peppers from a supermarket shelf, moving to place them into Tian Yi’s basket, and finally, Tian Yi transporting the fully loaded basket to the checkout counter for payment.
Each action step is paired with a corresponding image frame and labeled with action types like \textit{``Self''},  \textit{``Others''}, and \textit{``Move''}, providing clear multimodal grounding for hierarchical vision-language-action learning.
The annotation results show that our annotation scheme can accurately segment the key actions in the video and provide precise language descriptions of these key actions.
These detailed annotations decompose long-horizon tasks into simple, short-horizon robotic operations, each paired with precise natural language descriptions. This structured supervision enables VLA (Vision-Language-Action) models—such as RT-1~\cite{brohan2022rt}, OpenVLA~\cite{kim2024openvla}, RDT-1B~\cite{liu2024rdt}, $\pi_0$~\cite{black2024pi_0}, $\pi_{0.5}$~\cite{black2025pi_0.5}, and XR-1~\cite{fan2025xr}—to learn fundamental manipulation skills by aligning perceptual inputs, language instructions, and low-level actions in a scalable and interpretable manner.

Through standardized data collection, quality inspection, and annotation pipelines, RoboMIND 2.0 achieves multi-dimensional diversity and high reliability. Below, we systematically analyze the dataset’s characteristics and advantages across five core dimensions (morphology, tasks, objects, information, and simulation ) in Section~\ref{sec:dataset} and experimentally validate its effectiveness in imitation learning, VLA model training, and related applications in Section~\ref{exp:robomindv2}.

\section{\ours Dataset Analysis}
\label{sec:dataset}

While many large-scale robot manipulation datasets~\cite{khazatsky2024droid,o2024open,wu2025robocoin} claim “diversity”, the true value of diversity lies in which dimensions are covered and how well they reflect real-world variation. Different axes, such as tasks, objects, scenes, viewpoints, and interaction locations, influence a model’s generalization capabilities in distinct ways. To rigorously assess \ours, we evaluate its diversity across five foundational axes (\textit{Embodiment}, \textit{Task}, \textit{Object}, \textit{Information}, and \textit{Simulation}).

 \subsection{Embodiment Diversity}


Embodiment diversity is a core feature of RoboMIND 2.0, encompassing six heterogeneous dual-arm robotic platforms that exhibit rich variations across four key dimensions: kinematic structure, mobility capability, perception layout, and teleoperation modality. In contrast to existing large-scale manipulation datasets which predominantly rely on single-arm systems or homogeneous robot morphologies, we employ a unified cross-morphology data collection pipeline that ensures consistency while fully capturing the heterogeneity of robot embodiments.

\paragraph{Kinematic structure diversity.}
We incorporate both symmetric and asymmetric dual-arm configurations. The Franka and UR5e robots are arranged in a parallel setup (see Figure~\ref{fig:htact}), forming a kinematically symmetric layout with nearly identical reachable workspaces and joint topologies across the two arms. This configuration provides clean supervision for learning bilateral coordination patterns such as object handover, symmetric pushing, and cooperative grasping.
In contrast, the humanoid-style UR5e configuration, where the two arms are mounted with different shoulder offsets, joint zero-positions, and workspace geometries, introduces explicit structural asymmetries that challenge cross-arm generalization. The Tien~Kung and Tian~Yi robots (see Figure~\ref{fig:tianyi}) further extend this axis with a full humanoid upper-body morphology while still using parallel grippers as end-effectors, featuring human-like shoulder placements and distinct link geometries.

\paragraph{Mobility and workspace diversity.}
Our dataset spans both static dual-arm manipulation and mobile dual-arm operation. 
Static embodiments (Franka, UR5e, and the fixed mode Tien Kung) operate within confined tabletop 
workspaces, enabling high-precision manipulation and fine-grained bimanual coordination. 
In contrast, the AgileX Cobot Magic, ARX, and Tian Yi robots integrate dual arms with a mobile base, 
introducing large-scale, dynamic, and spatially varied interactions. The robot–object relationship 
evolves continuously as the base moves, producing trajectories that couple long-horizon navigation 
with manipulation in kitchens, supermarkets, and industrial settings. This axis substantially 
expands the embodied space explored by the robot and provides data crucial for learning 
mobile-manipulation.


\paragraph{Teleoperation modality diversity.}
To further diversify demonstrations, we employ three complementary human-in-the-loop control 
interfaces. (i) The HACTS master--slave system~\cite{xu2025hacts} is used for Franka, UR5e, and Tien Kung, enabling 
high-fidelity bilateral control with precise joint-space mapping. (ii) A VR-based teleoperation 
pipeline is used for ARX, where upper-body arm motions are captured via head-mounted displays 
and the mobile base is commanded through VR controllers, supporting large-workspace bimanual 
tasks. (iii) Physical guiding is used for AgileX and Tian Yi, where linear and angular velocities of 
the mobile base are recorded as operators directly push the robot, yielding highly naturalistic 
navigation behaviors. Together, these complementary modalities produce multi-style 
demonstrations that capture variations in human control strategies, temporal rhythms, and 
interaction preferences.

The combination of heterogeneous kinematic structures, varying mobility capabilities, diverse 
sensor placements, and multi-modal teleoperation interfaces yields a dataset with unprecedented 
embodiment coverage. This diversity is essential for training general-purpose manipulation 
policies capable of transferring across robot morphologies, adapting to inconsistent sensory 
observations, and performing long-horizon tasks in both static and mobile settings. 
Such systematic embodiment variation provides a principled foundation for research in cross-embodiment generalization, morphology-conditioned policy learning, and scalable embodied intelligence. The diversity of robot embodiments provides essential hardware support for broad coverage of tasks and objects. Different robot embodiments are tailored to distinct task requirements across varied scenarios, thereby enabling the collection of rich, diverse interaction data between tasks and objects.

 \subsection{Task Diversity}

\begin{figure}[h]
    \centering
    \includegraphics[width=0.98\textwidth]{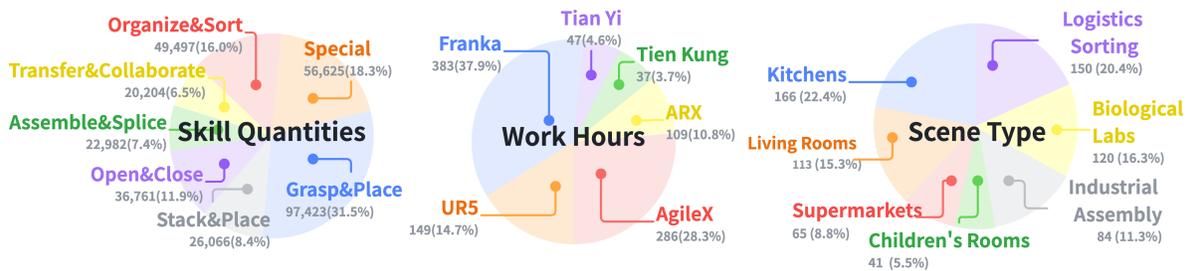}
    \caption{\textbf{Multi-metric distribution of RoboMIND 2.0.} \textbf{Skill Quantities:} Covers 7 core manipulation skills, with ``Grasp and  Place'' (31.5\%, 97,423 trajectories) as the most frequent skill, followed by ``Special'' (18.3\%, 56,625 trajectories), reflecting the dataset’s coverage of both basic and complex operations. \textbf{Robot Platform Work Hours:} Statistics of data collection hours across 6 robot embodiments, where Franka (37.9\%) and AgileX (28.3\%) contribute the most, aligning with their roles in fixed-scene and mobile manipulation tasks respectively. \textbf{Scene Type Proportions:} Includes 4 domestic scenes (52.0\%) and 3 industrial scenes (48.0\%), with ``Kitchens'' (22.4\%) and ``Logistics Sorting'' (20.4\%) as the dominant scenes in each category, ensuring scenario diversity for generalizable policy training.}
    \label{fig:task}
\end{figure}

\ours exhibits a highly structured and multi-dimensional task diversity that spans 
a total of 759 dual-arm manipulation tasks, systematically organized across two primary 
interaction settings: fixed-scene manipulation and mobile-scene manipulation. All tasks involve 
bimanual coordination, yet the spatial structure of the environment, whether the robot operates at 
a stationary workstation or navigates through larger, unstructured spaces, introduces fundamentally 
different forms of perception–action coupling and task complexity. Fixed-scene tasks emphasize precision object interaction, tabletop coordination, 
and structured spatial constraints, while mobile-scene tasks require dual-arm manipulation 
interleaved with spatial relocation, long-horizon navigation, and dynamic viewpoint adaptation.

A central dimension of task diversity comes from  manipulation skills. 
Across the entire dataset, tasks decompose into seven major skill families—grasping and placing, 
special-action manipulations, organizing and sorting behaviors, transfer and collaboration, 
assembly and splicing, opening–closing operations, and stacking–placement routines. Their empirical 
distribution reveals the functional breadth of the dataset. As is shown in Figure~\ref{fig:task}, grasping and placing constitutes the 
largest portion (97K+ episodes), followed by special-action tasks (56K+), organizing–sorting 
behaviors (49K+), opening–closing operations (36K+), stacking–placement tasks (26K+), 
assembly–splicing behaviors (22K+), and transfer–collaboration tasks (20K+). The coexistence of 
high-frequency foundational skills and low-frequency, structurally intricate skills ensure that 
manipulation policies must learn both short-horizon primitives and long-horizon, multi-step 
coordination strategies.

Task diversity is further enriched by the cross-embodiment distribution of tasks. Different 
robots contribute distinct task families due to their morphology, workspace geometry, embodiment 
constraints, and preferred deployment scenarios (see Figure~\ref{fig:task}). Franka, for instance, contributes the largest 
portion of tasks (224 tasks, 153K+ episodes) and accumulates 383 working hours, reflecting its 
strength in high-precision, fine-grained tabletop bimanual manipulation. UR5 contributes 162 tasks 
and 149 hours, characterized by wider-reach bimanual motions and extended workspace coverage. 
AgileX contributes 143 tasks and 286 hours across both fixed and mobile scenarios, demonstrating its 
role in large-scale, navigationally dependent dual-arm tasks.
ARX contributes 75 tasks with 109 hours, often involving VR-mediated demonstrations for spatially extended behaviors. Tien Kung
and Tian Yi, though contributing smaller numbers of tasks (49 and 46 respectively), introduce 
human-inspired upper-body kinematics and mobile-base dual-arm coordination, adding morphological 
variety and task types that require articulated reasoning or cross-room collaboration.

The task diversity of RoboMIND 2.0 is further amplified by its balanced coverage across scenes, as shown in the right pie chart of Figure~\ref{fig:task}. The dataset spans seven representative scenarios, evenly distributed between domestic (52.0\%) and industrial (48.0\%) settings, each characterized by distinct manipulation demands.
In domestic environments, \textbf{Kitchens} (22.4\%, 166 tasks) serve as the primary setting for everyday operations such as food handling and kitchenware manipulation. \textbf{Living Rooms} (15.3\%, 113 tasks), \textbf{Supermarkets} (8.8\%, 65 tasks), and \textbf{Children’s Rooms} (5.5\%, 41 tasks) respectively support basic human-object interactions, shopping-related activities, and fine-grained manipulation tasks.
In industrial contexts, \textbf{Logistics Sorting} (20.4\%, 150 tasks) focuses on large-scale, high-throughput workflows; \textbf{Biological Labs} (16.3\%, 120 tasks) emphasize high-precision, low-tolerance procedures; and \textbf{Industrial Assembly} (11.3\%, 84 tasks) involve complex, multi-step collaborative manipulations.
This deliberate alignment between scenes and task semantics ensures that RoboMIND 2.0 not only covers a broad spectrum of real-world manipulation challenges, from daily living to industrial automation, but also provides rich contextual variation for training policies that generalize across application domains.

Overall, the task diversity of \ours emerges from the interplay between skill variety, 
environmental structure, robot embodiment, and temporal scale. 
Such diversity provides a powerful foundation for training embodied agents that must adapt across tasks, scenes, and morphologies, and supports rigorous research into cross-task generalization, 
long-horizon reasoning, and robust open-world manipulation.

 \subsection{Object Diversity}

\begin{figure}[h]
    \centering
    \includegraphics[width=0.98\textwidth]{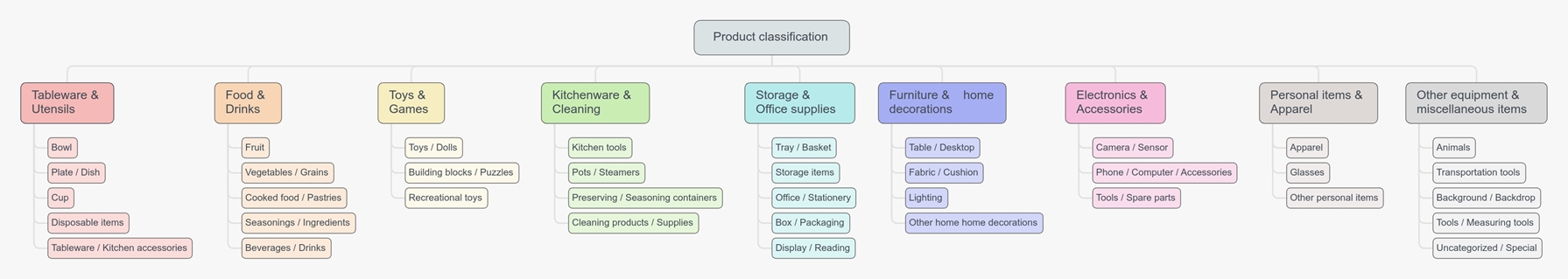}
    \caption{\textbf{Object category in \ours.} A hierarchical taxonomy of manipulation object categories in the dataset, organized into nine
    main groups with detailed subcategories listed under each, providing a structured overview of the 1,139 distinct objects used in manipulation tasks.}
    \label{fig:object}
\end{figure}

 Our dataset contains manipulation tasks involving 1,139 distinct objects, ranking it as the largest uniformly collected public dataset for robot manipulation in terms of the number of interacted objects to date.
 Figure~\ref{fig:object} illustrates the comprehensive product category taxonomy underpinning our dataset, which encompasses objects meticulously organized into six major categories: Food, Daily Necessities, Kitchenware, Stationery, Toys, and Others. Each category is further subdivided into fine-grained subcategories (e.g., fruits, snacks, cleaning supplies, cookware, writing instruments, educational toys, etc.). The breadth and granularity of this object inventory reflect an unprecedented level of real-world coverage, spanning everyday household items to specialized industrial tools, and including significant intra-category visual and geometric variation (e.g., multiple types of bottles, containers, or utensils). 

Our dataset exhibits well-structured and extensive object diversity, providing critical support for the model’s generalization to unseen objects and novel interaction scenarios. 
Rather than being limited to tabletop items, the dataset includes objects 
sourced from a wide range of real-world environments, including household tabletops, kitchen 
workspaces, children’s rooms, supermarket checkout and shelving areas, research laboratories, 
logistics and packing stations, storage environments, and door-centric interaction regions. The variation across these scenes in visual appearance, spatial layout, functional role, and context further amplifies the heterogeneity of object-task relationships.

A distinctive aspect of this object diversity lies not only in the visual or physical attributes of 
the objects, but also in the variety of manipulation types through which they are engaged. 
Across the dataset, objects participate in seven major manipulation categories: grasping and 
placing, special-action manipulations, organizing and sorting behaviors, transfer and 
collaboration tasks, assembly and splicing behaviors, opening–closing operations, and stacking 
and placement routines. These categories collectively span a spectrum of interaction intents, 
ranging from elementary pick-and-place skills to multi-stage reorganization, collaborative 
transportation, articulated-object interactions, and structured assembly processes.

Because most objects appear under multiple manipulation types rather than a single stereotyped 
pattern, the dataset captures the combinatorial and context-dependent nature of real-world 
interaction. The same object may be grasped on a tabletop, sorted inside a children’s room, carried 
across rooms in a household environment, assembled with other components in a laboratory setup, 
stacked during organization tasks in storage or logistics stations, or used within door-related 
operations that require opening, closing, or spatial transition. This rich cross-context reuse leads 
to a high-dimensional joint distribution over object attributes, environmental settings, and 
manipulation behaviors.

Overall, the combination of diverse objects, multi-scene deployment, and the full spectrum of 
seven structured manipulation categories provides a principled foundation for studying 
cross-object generalization, context-conditioned manipulation policies, and robust open-world 
performance in embodied agents.
\subsection{Information Diversity}

\begin{wrapfigure}{r}{0.4\textwidth}
\vspace{-1em}
  \centering
\includegraphics[width=0.35\textwidth]{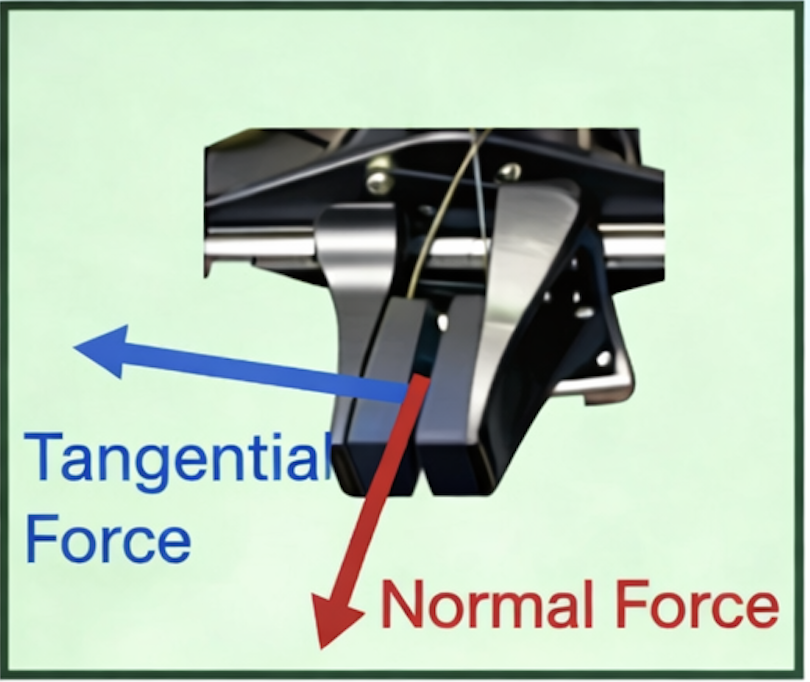}
  \caption{
    Tactile sensors with force components on grippers.
  }
  \label{fig:tactile}
  \vspace{-2em}
\end{wrapfigure}

Currently, most publicly available datasets provide only visual observations and proprioceptive (robotic embodiment) states, primarily for training imitation learning and reinforcement learning algorithms. However, as research in robot learning has advanced, researchers  increasingly incorporate tactile modalities into robot learning frameworks~\cite{yu2025forcevla,he2025foar,liu2025forcemimic}, significantly improving success rates on real-world robotic manipulation tasks.
In \ours, we not only record visual observations and robot proprioceptive states during task execution, but also use the Tashan Tactile sensors~\cite{tashantec2025} to collect tactile information as the robot performs long-horizon mobile manipulation tasks.

The tactile sensors record the tangential force, the normal force, and the direction of the normal force exerted on objects during the robotic arm’s task execution. Figure~\ref{fig:tactile} illustrates the end-effector setup on an AgileX mobile manipulator, featuring a parallel-jaw gripper integrated with two high-resolution tactile sensors, one on each finger. Each tactile sensor comprises two independent sensing modules, and each module captures three key physical quantities in real time: the normal force (perpendicular to the contact surface), the tangential force (parallel to the surface), and the direction of the tangential force (encoded as an angular orientation within the sensor plane). This rich multimodal tactile feedback enables fine-grained perception of contact geometry, slip detection, and interaction dynamics, which are critical for dexterous manipulation in unstructured environments.

\subsection{Simulation Diversity}
Unlike existing open-source robot datasets that release only real-robot datasets~\cite{o2024open,khazatsky2024droid,jiang2025galaxea,wu2025robocoin}, our dataset additionally provides the simulation assets of all objects used in the dataset, along with their integration into Isaac Sim to enable simulated data collection. Following ArtVIP~\cite{jin2025artvip},
we hire professional 3D modelers to create the simulation assets following a unified standard. 
Specifically, each articulated object in our dataset is structured into three levels: Assembly – Module – Mesh. We define a unified coordinate system at the base center and assemble parts bottom-up, integrating joints for dynamic motion and adding pixel-level affordance labels to mark interactive regions.
For visual realism, we use high-quality manifold meshes with smooth surfaces and high-resolution textures aligned to UV maps. Objects are rendered in Isaac Sim using RTX-based physically based rendering (PBR) for realistic material appearance.
For physical accuracy, we combine convex hulls, convex decomposition, and detailed collision meshes. We also improve Isaac Sim’s joint dynamics by adding position-dependent stiffness and velocity-dependent friction, validated with 0.1 mm precision optical motion capture.

We then import these assets into the Isaac Sim simulator and use them to collect 20K simulated trajectories on the dual-arm Franka and Tien Kung robots, performing the same tasks as in the real-world dataset.
In our experiments,  we train imitation learning models on a hybrid dataset combining real and simulated trajectories, which consistently improves real-world robot performance compared to training on real data alone. This not only validates the high fidelity of our digital twin, but also demonstrates that low-cost simulation data can effectively augment scarce real-world demonstrations, enhancing policy robustness and generalization. Thus, our work highlights simulation not just as a convenient proxy, but as an essential, cost-efficient engine for scalable robot learning.

\section{The MIND-2 Dual-System Model}

We propose a robotic operating system, consisting of a VLM-based “robot brain" slow system (MIND-2-VLM) and a VLA-based fast system for robotic manipulation (MIND-2-VLA).
The robotic brain slow system MIND-2-VLM is a Vision-Language Model that generates subtask instructions for the VLA model MIND-2-VLA, which executes actions based on the current task state and dynamic changes in the physical environment.
The robot execution fast system MIND-2-VLA is a Vision-Language-Action (VLA) model that generates control commands for the robot's base and dual-arm manipulation based on the task instructions, the robot's current state, and visual information of the task environment.

For the MIND-2-VLM model, we fine-tune the open-source VLM model InternVL3-8B~\cite{zhu2025internvl3} using the segmented subtask dataset from the \ours mobile robotics dataset, enabling it to determine which subtask the robot should execute based on its current state.
During the process of collecting our robot dataset, we generate a large amount of task failure data.
However, these failure data are not meaningless; instead, they provide valuable information that can be leveraged to help the robot model avoid such erroneous behaviors. Thus, we adopt an offline reinforcement learning approach to train the MIND-2-VLA model using both failure and success trajectories, enabling it to learn not only optimal actions from successful experiences but also to recognize and avoid suboptimal or incorrect behaviors demonstrated in failure cases.

\subsection{MIND-2-VLM: Building a Brain-Inspired VLM}

\begin{wrapfigure}
{r}{0.6\textwidth}
  \centering
\includegraphics[width=0.6\textwidth]{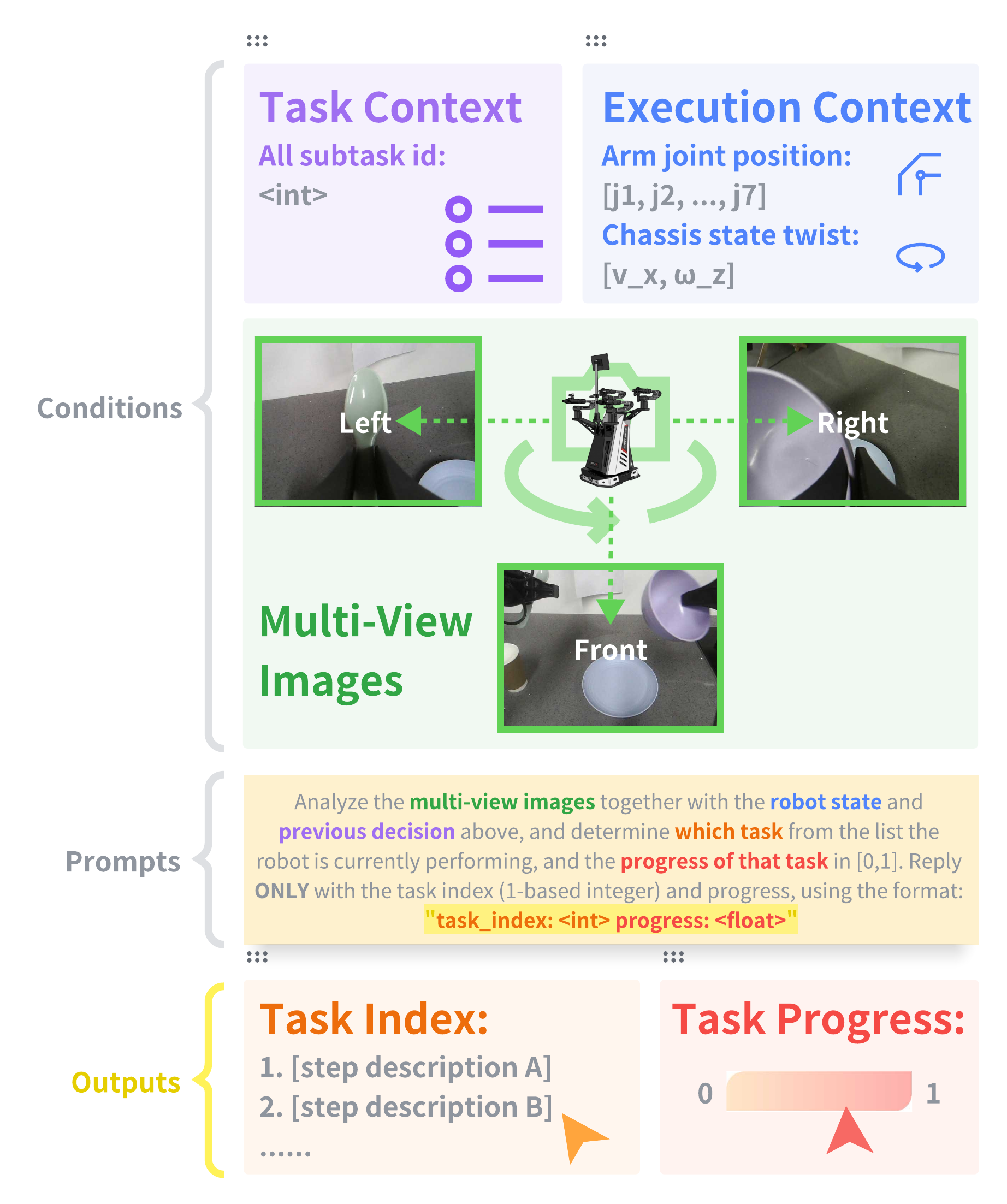}
  \caption{
    Prompt design for temporal task localization.
  }
  \label{fig:rlvlm}
\end{wrapfigure}

To enable high-level task reasoning from multimodal observations, we fine-tune a open-source vision-language model InternVL3-8B~\cite{zhu2025internvl3} on the \ours  that maps robot-centric visual and proprioceptive inputs to structured semantic outputs. Each training sample is constructed as a single-turn instruction-following dialogue. Figure~\ref{fig:rlvlm} illustrates how the collected dataset is used to fine-tune InternVL3-8B for robotic task understanding and planning.  The input prompt which serves as the ``human'' utterance, integrates three key components: \textbf{Multiview visual context}: Three \texttt{<image>} tokens representing synchronized front, left wrist, and right wrist camera views at the current timestep. \textbf{Task context}: An enumerated list of all subtasks in the episode (e.g., ``1. Pick up the red block'', ``2. Place it into the bin''), derived from human or automatic temporal segmentation annotations. \textbf{Execution context}: Current robot state, including 7-DoF arm joint positions and chassis twist velocities; The previously inferred task index (or ``None'' for the initial frame), enabling rudimentary sequential awareness.

The model's output, which serves as the ``assistant'' response, is supervised to be a deterministic string of the form:
\begin{center}
\texttt{**Answer:** Task Index:  xx \\Task Progress: xx}
\end{center}
where the ground-truth task index is obtained by locating the current frame within the annotated task segments, and the progress is linearly interpolated between the segment’s start and end frames as
\[
\text{progress} = \frac{t - t_{\text{start}}}{t_{\text{end}} - t_{\text{start}}} \in [0, 1].
\]

During fine-tuning, \texttt{InternVL3-8B} is trained end-to-end to minimize the cross-entropy loss over this textual output, conditioned on the multimodal input. The strict output format ensures that predictions are directly parseable by the downstream policy system
This design transforms raw robot logs into a scalable source of language-conditioned supervision, enabling the slow reasoning module (MIND-2-VLM) to reliably interpret real-time observations within the context of long-horizon tasks.

\subsection{MIND-2-VLA: Offline Training of VLA with Implicit Q-Learning}
\label{sec:training_iql}

To train the fast robot execution policy MIND-2-VLA from large-scale, real-world robotic demonstration data without online interaction, we adopt \textbf{Implicit Q-Learning (IQL)}~\cite{kostrikov2021offline}, an off-policy offline reinforcement learning algorithm well-suited for heterogeneous datasets containing both successful and suboptimal trajectories. IQL decouples Value and Q-function learning from policy optimization, enabling effective learning from mixed-quality data.

We begin by annotating all trajectories—both positive (successful) and negative (failed)—with temporally discounted reward signals to provide dense supervision. Specifically, For \textbf{successful trajectories}, the terminal step is assigned a reward of $+1$, and each preceding $t$  step receives the reward of  $r_t = \gamma^{T-t}$ with $\gamma = 0.999$, forming a decaying positive return-to-go. For \textbf{failure trajectories}, the terminal step is assigned $-1$, and earlier $t$ steps of reward are discounted backward as $r_t = -\gamma^{T-t}$, ensuring that actions leading to failure are associated with increasingly negative returns. $T$  denotes the number of time steps (or frames) in the action sequence—i.e., the length of the trajectory or the total number of sequential robot actions and observations in a given episode.

After defining the reward for the action sequence, we then introduce how to leverage Implicit Q-Learning (IQL) to perform offline reinforcement learning (RL) fine-tuning of Vision-Language-Action (VLA) models. 
IQL operates solely on a fixed dataset $\mathcal{D} = \{(s, a, r, s')\}$ collected by an unknown behavior policy and consists of three components: a state value function $V_\phi(s)$ (parameterized by $\phi$), a Q-function $Q_\theta(s, a)$ (parameterized by $\theta$), and a policy $\pi_\psi(a \mid s)$ (parameterized by $\psi$).
We first learn a state value function $V_\phi(s)$ by regressing it towards the expected return under the behavior policy in the dataset $\mathcal{D}$. This is achieved by minimizing the following loss:
\begin{equation}
    \mathcal{L}_V(\phi) = \mathbb{E}_{(s,a) \sim \mathcal{D}} \left[ \rho_\tau \left( Q_{\text{target}}(s, a) - V_\phi(s) \right) \right],
    \label{eq:iql_v_loss}
\end{equation}
\noindent where $Q_{\text{target}}$ denotes a target Q-network (e.g., an exponential moving average of $Q_\theta$), and $\rho_\tau(u)$ is the asymmetric quantile loss given by
\begin{equation}
    \rho_\tau(u) = u \cdot \big( \tau - \mathbb{I}(u < 0) \big),
    \label{eq:quantile_loss}
\end{equation}
where $\mathbb{I}(\cdot)$ denotes the indicator function, and $\tau \in (0,1)$ is the target quantile level. 

Next, we update the Q-function $Q_\theta(s,a)$ using standard temporal difference (TD) learning with target values computed from $V(s')$:
\begin{equation}
    \mathcal{L}_Q(\theta) = \mathbb{E}_{(s,a,s') \sim \mathcal{D}} \left[ \frac{1}{2} \left( Q_\theta(s,a) - (r(s,a) + \gamma^{rl} V(s')) \right)^2 \right],
\end{equation}
where $r(s,a)$ is the observed scalar reward, and $\gamma^{rl} \in [0,1)$ is the discount factor, and the expectation is taken over full transitions sampled from $\mathcal{D}$.

Finally, the policy $\pi_\psi(a|s)$ is updated via advantage-weighted regression, focusing on actions that outperform the average value in each state. The policy loss is:
\begin{equation}
    \mathcal{L}_\pi(\psi) = -\mathbb{E}_{(s,a) \sim \mathcal{D}} \left[ \exp(\beta \cdot A(s,a)) \cdot \log \pi_\psi(a|s) \right],
\end{equation}
where $A(s,a) = Q(s,a) - V(s)$ is the implicit advantage, and $\beta > 0$ is a temperature hyperparameter.

The inclusion of failure trajectories is critical: by contrasting high-return and low-return sequences, IQL learns to distinguish between actions that appear plausible but lead to failure (e.g., weak grasps or misalignments) and those that reliably succeed. This allows MIND-2-VLA to not only imitate successful behaviors but also actively avoid known failure modes—enabling robust, generalizable, and safe policy learning from real-world, open-world data.
\section{\ours Experiments}
\label{exp:robomindv2}
\begin{figure}[tb]
  \centering
  \includegraphics[width=1.0\columnwidth]{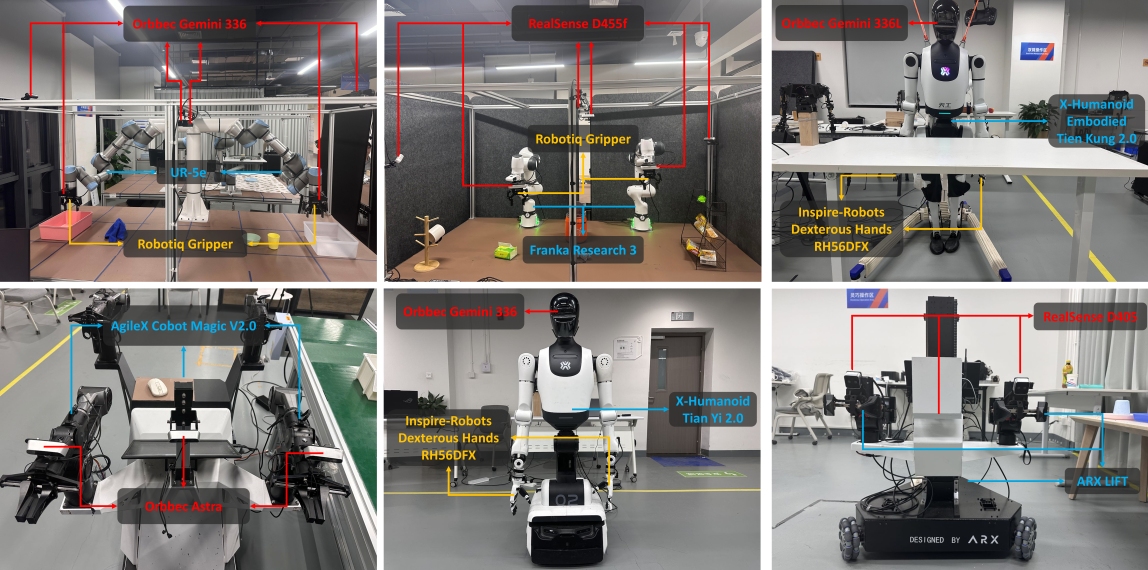}
  \caption{
    \textbf{Robotic real-world setup.}
    For the Franka and UR5e robots, we use cameras positioned at the top, left, and right viewpoints to record the visual information of the task trajectories.
For the humanoid (Tien Kung and Tian Yi) robots, we use their built-in RGB-D cameras to capture visual observations.
For the AgileX and ARX robots, we use dual wrist-mounted cameras (one on each arm) as well as a head-mounted camera to capture visual information.
  }
  \label{fig:hardware_setup}
\end{figure}

\begin{figure*}[tb]
  \centering
  \subfloat{\includegraphics[width=0.98\textwidth]{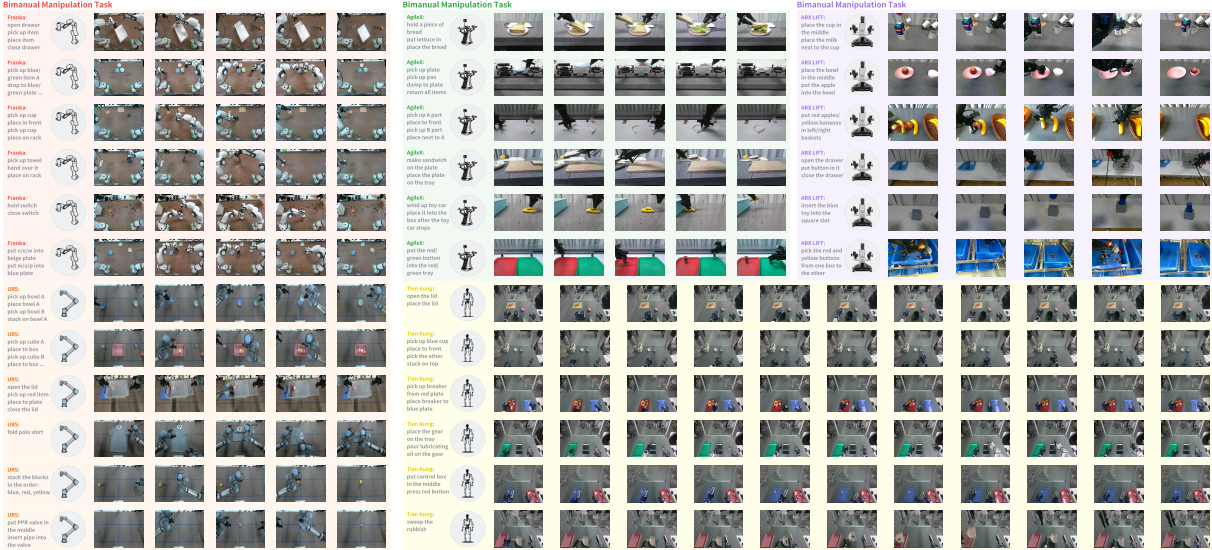}}
  \caption{
    \textbf{Diverse bimanual manipulation tasks across multiple robot platforms.}
    A variety of bimanual manipulation tasks are performed by different robotic platforms—including Franka, UR5e, AgileX, ARX, and Tien Kung, which are tested on different baselines.
  }
  \label{fig:il_tasks}
\end{figure*}

\begin{figure*}[h]
  \centering
  \subfloat{\includegraphics[width=0.98\textwidth]{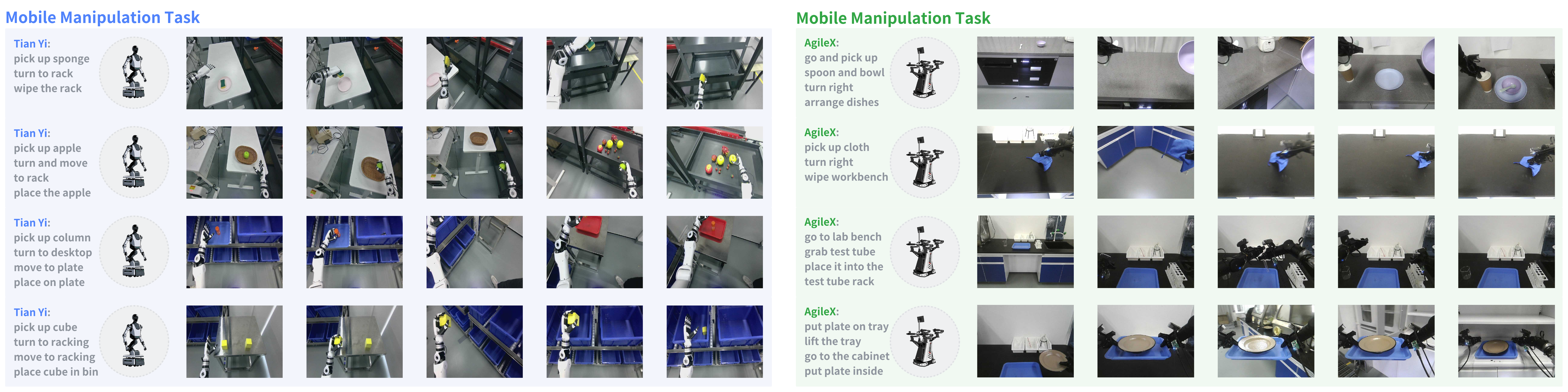}}
  \caption{
    Examples of mobile manipulation tasks performed by the Tian Yi and AgileX robot platforms in diverse real-world environments.
  }
  \label{fig:mob_tasks}
\end{figure*}

In this section, we present the experimental results evaluating the \ours dataset, including assessments of the open-source MIND-2 system, digital twin framework, and object generalization capabilities.


\ours serves as a benchmark to evaluate the capabilities and limitations of these approaches. We assess not only single-task imitation learning models, but also vision-language-action (VLA) large models  that can perform multiple tasks using \ours. Notably, \ours now includes a substantial amount of mobile manipulation data, enabling evaluation across both mobile and stationary (fixed-skill) scenarios. This dual evaluation provides a more holistic understanding of model generalization and robustness in diverse robotic settings.
We further evaluate MIND-2 through a series of experiments to assess its capabilities in both single robot and multiple robots settings. Specifically, we test its performance on single-robot mobile manipulation tasks using the AgileX platform, as well as on three newly designed multi-robot collaboration tasks that require coordinated long-horizon interaction. 
In addition, we investigate whether post-training with Implicit Quantile Learning (IQL) improves the policy’s action accuracy and robustness, and whether pretraining MIND-2 on our full-scale mobile manipulation dataset which contains long-horizon trajectories collected across diverse real-world environments leads to significant gains in downstream task success. Together, these experiments probe the impact of large-scale mobile data, multi-agent coordination, and value-based refinement on the effectiveness of VLA-based robotic policies. Finally, we conduct three targeted experiments to further analyze the capabilities of VLA models: (1) object generalization—evaluating whether the model can successfully manipulate novel objects with unseen shapes, colors, or materials; (2) the role of tactile feedback—assessing how integrating touch sensing improves manipulation robustness and success; and (3) sim-to-real transfer—investigating whether training on our high-fidelity simulator dataset enhances performance on real-robot tasks. 


\subsection{Imitation Learning and Vision-Language-Action Models Experiments}
\label{sec:exp-benchmark-overview}

This section provides a unified benchmark evaluation of two representative paradigms for real-world robot learning on \ours: \emph{single-task imitation learning methods} and \emph{vision-language-action (VLA) large methods}. 

By evaluating both paradigms under the same task settings, robot embodiments, and real-world execution conditions, we seek to systematically analyze their respective strengths and limitations in terms of task success, embodiment generalization, and robustness to long-horizon manipulation. 
In particular, the benchmark spans fixed-base dual-arm manipulation, mobile manipulation, and humanoid bimanual tasks, covering a wide spectrum of kinematic structures, sensing configurations, and interaction complexity. 
This unified evaluation not only enables a fair comparison between task-specialized policies and large-scale foundation models, but also sheds light on how data diversity, embodiment alignment, and action representation affect real-world robotic performance.

\subsubsection{Experiment Tasks}
\label{subsec:hardware_setup}

To evaluate VLA models and single-task imitation learning models on the \ours, we use three benchmark tasks: fixed-scene dual-arm manipulation, mobile dual-arm manipulation, and humanoid dual-arm manipulation.
Figures~\ref{fig:il_tasks} and~\ref{fig:mob_tasks} present visualizations of evaluation tasks from \ours, including fixed-base bimanual, humanoid dual-arm, and mobile dual-arm robots.

We evaluate a diverse set of dual-arm manipulation tasks across five robotic platforms: \textbf{Franka}, \textbf{UR5e}, \textbf{ARX}, \textbf{AgileX}, and \textbf{Tian Yi/Tien Kung}. For each platform, we specify whether the robot operates in a fixed-base or mobile configuration.

\paragraph{Franka (Fixed-base).}
\begin{itemize}[left=0pt]
    \item Hand a cup from the right arm to the left arm for hanging on a rack.
    \item Similarly pass and hang a towel.
    \item Stabilize a load switch with the left arm while the right flips it on.
    \item Open a cabinet compartment (left), retrieve a yellow button (right), and close the compartment (left).
    \item Sort 12 mixed lowercase letters: ``new'' into a beige tray, ``map'' into a blue tray; nothing into the green tray.
    \item Move only blue and green letters (out of 12 colored letters) to their matching trays; leave others in place and keep the beige tray empty.
\end{itemize}

\paragraph{UR5e (Fixed-base).}
\begin{itemize}[left=0pt]
    \item Place a blue bowl at the table center (left); insert a green bowl into it (right).
    \item Jointly grasp blocks and place them into a pink storage box.
    \item Open the storage box lid (right) while simultaneously retrieving a red block and placing it onto a blue tray (left)—requiring smooth, continuous motion without dropping.
    \item Collaboratively fold clothes.
    \item Stack blocks vertically at the table center in order: blue (bottom), red, yellow (top).
    \item Position a PPR valve at the center (left); align, insert, and tighten a connector tube into its port (right).
\end{itemize}

\paragraph{ARX (Fixed-base).}
\begin{itemize}[left=0pt]
    \item Place a cup at the table center and position a milk carton beside it.
    \item Transfer an apple from a plate into a bowl.
    \item Collaboratively sort fruits by color: red apples into the left basket, yellow bananas into the right.
    \item Pull out a shelf (right), place a button on it (left), and push it back in (right).
    \item Insert a blue part into a gray base.
    \item Select red and yellow buttons (left) and hand them to the right arm for placement into a box.
\end{itemize}

\paragraph{AgileX.}
The AgileX robots operate in both fixed-base and mobile configurations:
\begin{itemize}[left=0pt]
    \item\textit{Fixed-base}: 
    \begin{enumerate}[label=(\arabic*), nosep, left=0pt]
        \item Assemble a sandwich: place lettuce on bread held by the left arm (right).
        \item Lift a plate (right) and pour meat onto it.
        \item Cap the sandwich with a top bread slice (right); carry the plate to a tray (left).
        \item Place a red button into its color-matched tray (left).
        \item Place a green button into the corresponding green tray (right).
        \item Wind up a yellow Beetle toy car (right); once stopped, pick it up along with drainage pipe segments and place them into a blue bin (left).
    \end{enumerate}
    \item\textit{Mobile}:
    \begin{enumerate}[label=(\arabic*), nosep, left=0pt]
        \item Pick up a spoon and a bowl and arrange them as a set.
        \item Wipe the workbench using a cloth.
        \item Place a test tube into a rack.
        \item Transport a plate on a tray into a cabinet.
    \end{enumerate}
\end{itemize}

\paragraph{Tian Yi/Tien Kung}
The Tian Yi platform is used in a wheeled mobile configuration, while Tien Kung is a humanoid robot for dexterous bimanual tasks:
\begin{itemize}[left=0pt]
    \item \textit{Tian Yi (Mobile)}:
    \begin{enumerate}[label=(\arabic*), nosep, left=0pt]
        \item Wipe dust from the middle shelf with a cloth.
        \item Pick up a yellow foam cube from the table and place it on the middle level of the right shelf.
        \item Grasp an orange capacitor from the left shelf and place it onto the right tabletop.
        \item Retrieve a green apple from a basket on the table and place it on the middle level of the left shelf.
    \end{enumerate}
    \item\textit{Tien Kung (Humanoid, Fixed-base)}:
    \begin{enumerate}[label=(\arabic*), nosep, left=0pt]
        \item Open a pot lid.
        \item The left arm picks up a circuit breaker from a red tray and places it at the table center; the right arm then retrieves it and places it into a blue tray on the right.
        \item Pour lubricating oil onto a gear.
        \item Stack cups.
        \item The left arm places a control box at the table center while the right arm presses the red emergency stop button on it.
        \item Sweep electronic component debris into a waste bin.
    \end{enumerate}
\end{itemize}

We categorize the six fixed-scene tasks into three difficulty levels: Easy, Medium, and Hard. Easy-level tasks typically involve bimanual object passing with simple transfer motions. Medium-level tasks require bimanual coordination to perform fine-grained pushing, pulling, picking, and placing operations. Hard-level tasks involve long-horizon actions that require categorizing objects based on their spatial arrangements, or performing detailed assembly operations using both arms.

\subsubsection{Real-world Robotic Setup}
Our real-world robotic setup is shown in Figure~\ref{fig:hardware_setup}.
Each embodying distinct design philosophies and capabilities for bimanual manipulation. These include: (1) a parallel-mounted dual-arm UR5e system with Robotiq grippers~\cite{robotiq2026adaptive} and an Orbbec Gemini 336 stereo camera~\cite{orbbec2026}; (2) a dual-arm Franka  paired with a Robotiq gripper~\cite{robotiq2026adaptive} and a RealSense D455F depth camera~\cite{intel2026realsense}; (3) the humanoid X-Humanoid Tien Kung 2.0 equipped with dexterous RH56DFX hands~\cite{inspire2026rh56dfx} and an Orbbec Gemini 336L camera~\cite{orbbec2026}; (4) the mobile AgileX Cobot Magic V2.0 featuring three Orbbec Astra camerasa~\cite{orbbec2026}, which are mounted on the left arm, right arm, and the robot’s primary (front-facing) viewpoint; (5) the full-body humanoid Tian Yi 2.0, also with RH56DFX hands~\cite{inspire2026rh56dfx} and a central Orbbec Gemini 336 cameraa~\cite{orbbec2026}; and (6) the ARX  mobile manipulator integrated with a RealSense D435i camera~\cite{intel2026realsense} for navigation and object interaction. Together, these platforms span a broad spectrum of embodiment types, from fixed to mobile and from industrial arms to humanoids, providing rich variation in kinematics, workspace geometry, end-effectors, and visual sensing, which is essential for training and evaluating generalizable vision-language-action models in real-world settings.

\begin{table}[H]
  \centering
  \vspace{-0.4cm}
  \caption{
    Performance comparison of single task imitation learning methods and VLA models across different task categories.
    \colorbox[HTML]{E0F4FF}{\textbf{Color boxes}} indicate the best-performing model in each task category. In the following table, we also adhere to these rules.
  }
  \vspace{-0.4cm}
  \resizebox{0.73\columnwidth}{!}{
    \setlength{\tabcolsep}{5pt}
    \begin{tabular}{l *{8}{c}}
      \toprule
      & \textbf{ACT~\cite{zhao2023act}} & \textbf{Dense Policy~\cite{su2025dense}} & \textbf{DP3~\cite{ze20243d}} & \textbf{UVA~\cite{li2025uva}} & \textbf{$\pi_{0}$~\cite{black2024pi_0}} & \textbf{$\pi_{0.5}$~\cite{black2025pi_0.5}} & \textbf{HybridVLA~\cite{liu2025hybridvla}} & \textbf{XR-1~\cite{fan2025xr}} \\
      \midrule
      \multicolumn{9}{c}{\textbf{Franka-Robot Tasks (FR)}} \\
      \cmidrule(lr){1-9}
      \texttt{FR-Task1} & 0.0 & \colorbox[HTML]{E0F4FF}{\textbf{0.3}} & 0.2 & 0.1 & 0.1 & 0.2 & 0.1 & 0.0 \\
      \texttt{FR-Task2} & 0.0 & 0.2 & 0.2 & 0.2 & 0.2 & 0.3 & 0.1 & \colorbox[HTML]{E0F4FF}{\textbf{0.8}} \\
      \texttt{FR-Task3} & 0.0 & 0.1 & 0.2 & 0.1 & 0.1 & 0.2 & 0.1 & \colorbox[HTML]{E0F4FF}{\textbf{0.3}} \\
      \texttt{FR-Task4} & 0.0 & 0.1 & 0.4 & 0.0 & 0.0 & 0.0 & 0.0 & \colorbox[HTML]{E0F4FF}{\textbf{0.7}} \\
      \texttt{FR-Task5} & 0.0 & 0.0 & 0.0 & 0.0 & 0.0 & 0.0 & 0.0 & 0.0 \\
      \texttt{FR-Task6} & 0.0 & 0.0 & 0.0 & 0.0 & 0.0 & 0.0 & 0.0 & 0.0 \\
      \midrule
      \multicolumn{9}{c}{\textbf{UR5e-Robot Tasks (UR)}} \\
      \cmidrule(lr){1-9}
      \texttt{UR-Task1} & 0.3 & 0.3 & 0.5 & 0.3 & 0.6 & \colorbox[HTML]{E0F4FF}{\textbf{0.8}} & 0.5 & \colorbox[HTML]{E0F4FF}{\textbf{0.8}} \\
      \texttt{UR-Task2} & 0.3 & 0.3 & 0.0 & 0.4 & 0.4 & 0.4 & 0.3 & \colorbox[HTML]{E0F4FF}{\textbf{0.6}} \\
      \texttt{UR-Task3} & 0.5 & 0.4 & 0.3 & 0.5 & \colorbox[HTML]{E0F4FF}{\textbf{0.6}} & \colorbox[HTML]{E0F4FF}{\textbf{0.6}} & 0.5 & 0.5 \\
      \texttt{UR-Task4} & 0.1 & 0.2 & 0.3 & 0.3 & 0.4 & \colorbox[HTML]{E0F4FF}{\textbf{0.5}} & 0.3 & \colorbox[HTML]{E0F4FF}{\textbf{0.5}} \\
      \texttt{UR-Task5} & 0.4 & 0.4 & 0.5 & 0.4 & \colorbox[HTML]{E0F4FF}{\textbf{0.6}} & \colorbox[HTML]{E0F4FF}{\textbf{0.6}} & 0.5 & 0.4 \\
      \texttt{UR-Task6} & 0.2 & \colorbox[HTML]{E0F4FF}{\textbf{0.4}} & \colorbox[HTML]{E0F4FF}{\textbf{0.4}} & 0.2 & 0.2 & 0.3 & 0.2 & 0.2 \\
      \midrule
      \multicolumn{9}{c}{\textbf{AgileX Fixed-base Manipulation Tasks }} \\
      \cmidrule(lr){1-9}
      \texttt{AgileX-Task1} & 0.2 & 0.3 & 0.0 & 0.0 & 0.5 & \colorbox[HTML]{E0F4FF}{\textbf{0.6}} & 0.4 & \colorbox[HTML]{E0F4FF}{\textbf{0.6}} \\
      \texttt{AgileX-Task2} & 0.2 & 0.3 & 0.2 & \colorbox[HTML]{E0F4FF}{\textbf{0.4}} & 0.2 & 0.2 & 0.2 & \colorbox[HTML]{E0F4FF}{\textbf{0.4}} \\
      \texttt{AgileX-Task3} & 0.1 & 0.2 & 0.0 & 0.0 & 0.5 & \colorbox[HTML]{E0F4FF}{\textbf{0.6}} & 0.5 & \colorbox[HTML]{E0F4FF}{\textbf{0.6}} \\
      \texttt{AgileX-Task4} & 0.0 & 0.2 & 0.1 & 0.0 & 0.2 & 0.4 & 0.3 & \colorbox[HTML]{E0F4FF}{\textbf{0.5}} \\
      \texttt{AgileX-Task5} & 0.0 & 0.0 & 0.0 & 0.1 & 0.0 & 0.0 & 0.0 & \colorbox[HTML]{E0F4FF}{\textbf{0.2}} \\
      \texttt{AgileX-Task6} & 0.0 & 0.0 & 0.2 & 0.0 & 0.2 & 0.3 & 0.2 & \colorbox[HTML]{E0F4FF}{\textbf{0.4}} \\
      \midrule
      \multicolumn{9}{c}{\textbf{ARX Fixed-base Manipulation Tasks}} \\
      \cmidrule(lr){1-9}
      \texttt{ARX-Task1} & 0.1 & 0.2 & 0.3 & 0.0 & 0.1 & 0.2 & 0.1 & \colorbox[HTML]{E0F4FF}{\textbf{0.2}} \\
      \texttt{ARX-Task2} & 0.2 & 0.3 & 0.3 & 0.4 & 0.2 & 0.3 & 0.3 & \colorbox[HTML]{E0F4FF}{\textbf{0.3}} \\
      \texttt{ARX-Task3} & 0.0 & 0.0 & 0.0 & \colorbox[HTML]{E0F4FF}{\textbf{0.1}} & 0.0 & \colorbox[HTML]{E0F4FF}{\textbf{0.1}} & 0.0 & \colorbox[HTML]{E0F4FF}{\textbf{0.1}} \\
      \texttt{ARX-Task4} & 0.1 & 0.1 & 0.0 & 0.1 & 0.1 & 0.2 & 0.1 & \colorbox[HTML]{E0F4FF}{\textbf{0.3}} \\
      \texttt{ARX-Task5} & 0.1 & 0.1 & 0.2 & 0.1 & 0.1 & 0.2 & 0.1 & \colorbox[HTML]{E0F4FF}{\textbf{0.3}} \\
      \texttt{ARX-Task6} & 0.0 & 0.0 & 0.1 & \colorbox[HTML]{E0F4FF}{\textbf{0.2}} & 0.0 & 0.1 & 0.0 & \colorbox[HTML]{E0F4FF}{\textbf{0.2}} \\
      \midrule
      \multicolumn{9}{c}{\textbf{Tien Kung Fixed-base Manipulation Tasks}} \\
      \cmidrule(lr){1-9}
      \texttt{Tien Kung-Task1} & 0.5 & 0.3 & 0.0 & 0.2 & 0.3 & 0.4 & 0.3 & \colorbox[HTML]{E0F4FF}{\textbf{0.8}} \\
      \texttt{Tien Kung-Task2} & 0.2 & 0.1 & 0.1 & 0.1 & 0.1 & 0.5 & 0.2 & \colorbox[HTML]{E0F4FF}{\textbf{0.6}} \\
      \texttt{Tien Kung-Task3} & 0.0 & 0.0 & 0.0 & 0.0 & 0.0 & 0.3 & 0.1 & \colorbox[HTML]{E0F4FF}{\textbf{0.6}} \\
      \texttt{Tien Kung-Task4} & 0.1 & 0.1 & 0.0 & 0.0 & 0.2 & 0.4 & 0.1 & \colorbox[HTML]{E0F4FF}{\textbf{0.7}} \\
      \texttt{Tien Kung-Task5} & 0.1 & 0.2 & 0.1 & 0.2 & 0.2 & 0.2 & 0.2 & \colorbox[HTML]{E0F4FF}{\textbf{0.7}} \\
      \texttt{Tien Kung-Task6} & 0.1 & 0.1 & 0.1 & 0.0 & 0.0 & 0.1 & 0.0 & \colorbox[HTML]{E0F4FF}{\textbf{0.3}} \\
      \midrule
      \multicolumn{9}{c}{\textbf{AgileX Mobile Manipulation Tasks}} \\
      \cmidrule(lr){1-9}
      \texttt{AgileX-MV-Task1} & 0.2 & 0.3 & 0.1 & 0.0 & 0.1 & 0.3 & 0.2 & \colorbox[HTML]{E0F4FF}{\textbf{0.4}} \\
      \texttt{AgileX-MV-Task2} & \colorbox[HTML]{E0F4FF}{\textbf{0.4}} & \colorbox[HTML]{E0F4FF}{\textbf{0.4}} & 0.1 & 0.3 & 0.0 & 0.0 & 0.0 & 0.2 \\
      \texttt{AgileX-MV-Task3} & \colorbox[HTML]{E0F4FF}{\textbf{0.4}} & \colorbox[HTML]{E0F4FF}{\textbf{0.4}} & 0.0 & \colorbox[HTML]{E0F4FF}{\textbf{0.4}} & 0.0 & 0.3 & 0.2 & \colorbox[HTML]{E0F4FF}{\textbf{0.4}} \\
      \texttt{AgileX-MV-Task4} & 0.0 & 0.1 & 0.0 & \colorbox[HTML]{E0F4FF}{\textbf{0.3}} & 0.1 & 0.1 & 0.1 & \colorbox[HTML]{E0F4FF}{\textbf{0.3}} \\
      \midrule
      \multicolumn{9}{c}{\textbf{Tian Yi Mobile Manipulation Tasks}} \\
      \cmidrule(lr){1-9}
      \texttt{Tian Yi-Task1} & 0.2 & 0.2 & 0.1 & 0.1 & 0.1 & 0.2 & 0.1 & \colorbox[HTML]{E0F4FF}{\textbf{0.3}} \\
      \texttt{Tian Yi-Task2} & 0.1 & 0.1 & 0.1 & 0.2 & 0.2 & 0.2 & 0.2 & \colorbox[HTML]{E0F4FF}{\textbf{0.4}} \\
      \texttt{Tian Yi-Task3} & 0.1 & 0.1 & 0.0 & 0.0 & 0.1 & 0.3 & 0.2 & \colorbox[HTML]{E0F4FF}{\textbf{0.3}} \\
      \texttt{Tian Yi-Task4} & 0.1 & 0.3 & 0.0 & 0.3 & 0.2 & 0.3 & 0.2 & \colorbox[HTML]{E0F4FF}{\textbf{0.5}} \\
      \bottomrule
    \end{tabular}
  }
  \label{tab:task_comparison}
\end{table}

\subsubsection{Single-task Imitation Learning Models}
\label{Sec:exp-single}

\textbf{Training and Evaluation Setup.}
For single-task imitation learning, we adopt four established baseline methods: ACT~\cite{zhao2023act}, Dense Policy~\cite{su2025dense},  DP3~\cite{ze20243d}, and UVA~\cite{li2025uva}.
We follow the default model configurations as specified in their original publications.
Using these methods, we train each single-task model from scratch on the respective dataset.
After training, the models are directly deployed in real-world environments for evaluation.
Model performance is assessed based on task success rate.
Each model is tested ten times, with testers recording whether each trial succeeded or failed, along with the reasons for any failures.
This systematic evaluation provides valuable insights for future improvements.

\textbf{Experimental Results.}
The performance of  ACT, Dense Policy, DP3, and UVA is summarized in Table~\ref{tab:task_comparison}, reporting success rates across 38 tasks and all collected data of robot types.
The results in the table show the success rates of various imitation learning methods, including DP3, Dense Policy, ACT, and UVA, across a diverse set of tasks from multiple robot platforms: Franka, UR5e, AgileX, ARX, Tien Kung, and Tian Yi. These tasks vary significantly in complexity, embodiment, and environment, enabling a comprehensive evaluation of generalization capability.
DP3 demonstrates strong performance on tasks involving fixed-arm robots with rich visual context, particularly in the FR and UR benchmarks. For example, it achieves high success rates when provided with dense action supervision and multi-view observations. However, its performance drops on mobile or humanoid platforms such as AgileX-MV and Tian Yi, where the state space includes dynamic navigation and full-body coordination. This suggests that DP3 benefits from structured, stationary environments but struggles with cross-morphology generalization due to limited adaptation to varying kinematics and sensorimotor modalities.
In contrast, Dense Policy exhibits more consistent performance across different robot embodiments and task types. It performs well not only on fixed-base robots, but also shows moderate success on mobile manipulation tasks. This indicates that Dense Policy’s design enables better transferability across heterogeneous hardware and task domains.
Notably, both models achieve lower success rates on long-horizon collaborative tasks, especially those requiring precise handover or spatial reasoning (e.g., Franka-Task6 and ARX-Task3), suggesting that current imitation learning approaches still face challenges in modeling fine-grained coordination and long-horizon planning.
While DP3 excels in structured, visually rich environments with known robot configurations, Dense Policy shows stronger generalization capabilities across diverse robotic platforms, making it more suitable for open-world, multi-robot manipulation scenarios. Overall, incorporating 3D point cloud inputs leads to greater performance gains in imitation learning compared to using 2D image inputs alone.


\subsubsection{Vision-Language-Action Large Models}
\label{EXP:MUL}

\textbf{Training and Evaluation Setup.}
We evaluated the performance of four models (HybridVLA~\cite{liu2025hybridvla}, $\pi_0$~\cite{black2024pi_0}, $\pi_{0.5}$~\cite{black2025pi_0.5}, and XR-1~\cite{fan2025xr}) fine-tuned by the demonstrations from \ours in completing various real-world tasks.
During the training of the VLA model, we fine-tune the publicly available pre-trained VLA models on each task, using the same pre-training configuration and hyperparameters as reported in their published code.

\textbf{Experimental Results.}
Table~\ref{tab:task_comparison} presents the success rates for various robot tasks performed using the three different VLA models.
The evaluation results across six robot platforms, including Franka (FR), UR5e (UR), AgileX, ARX, Tien Kung, and Tian Yi, reveal distinct strengths and limitations of current VLA models in real-world imitation learning.
$\pi_0$, trained primarily on internet-scale data with limited robotic fine-tuning, shows modest performance overall. It achieves moderate success rates on simple UR5e  tasks  but consistently fails on bimanual or mobile manipulation scenarios (e.g., near-zero success on AgileX-MV and Tian Yi tasks). This suggests that while $\pi_0$ possesses  a broad semantic understanding, it lacks the grounded motor priors necessary for precise physical interaction in diverse embodiments.
$\pi_{0.5}$, an improved variant with additional robotic fine-tuning and better action tokenization, demonstrates noticeable gains over $\pi_0$, particularly on fixed-base dual-arm tasks such as UR-Task1 (0.8) and AgileX-Task1 (0.6). However, its performance still degrades significantly on mobile or humanoid systems.
Both $\pi_0$ and HybridVLA use strong vision-language models, but they generate robot actions differently.
$\pi_0$ uses a flow matching action head. It predicts the full action in one step, which is fast and works well for short tasks.
HybridVLA, on the other hand, combines autoregressive prediction (for high-level commands) with a diffusion model (for precise, continuous motions like joint movements).
Despite these architectural differences, our evaluation across six diverse robot platforms reveals that HybridVLA and $\pi_0$ achieve remarkably similar task success rates.
This observation suggests that, under comparable training data and vision-language encoders, the choice of action generation head alone, whether flow matching, diffusion, or autoregressive, does not lead to substantial gains in real-world task accuracy. Instead, factors such as data diversity, embodiment alignment, and long-horizon supervision may play a more decisive role in determining overall performance.
Among all evaluated VLA models, XR-1 demonstrates the strongest overall performance and remarkable cross-embodiment generalization. It consistently achieves high success rates across a wide range of robotic platforms, including fixed dual-arm systems, mobile manipulators, and full humanoid robots, where many other models struggle due to morphological and sensory heterogeneity. Unlike $\pi_0$ and $\pi_{0.5}$, which rely heavily on internet-scale pretraining but lack grounded motor priors, XR-1 leverages a tightly integrated architecture that fuses multimodal perception, proprioceptive feedback, and language-conditioned action generation. This enables it to accurately interpret task instructions and execute complex, coordinated behaviors, even in bimanual or dynamic environments.

\subsection{MIND-2 Dual System Experiments}
\label{exp:dualsystem}
In this section, we present the experimental evaluation of our MIND-2 fast-slow system. We find that for long-horizon mobile manipulation tasks, both mainstream imitation learning methods and existing Vision-Language-Action (VLA) models perform poorly. To validate the effectiveness of our proposed MIND-2 fast-slow system, we first evaluate it on mobile manipulation tasks using the AgileX robot. 
Table~\ref{tab:performance} shows that the MIND-2 fast-slow system achieves significantly better performance across various tasks compared to both VLA models and single-task imitation learning methods.

\begin{table}[htbp]
\centering
\caption{Performance comparison across AgileX  mobile manipulation tasks.}
\label{tab:performance}
\resizebox{0.85\columnwidth}{!}{
\begin{tabular}{l|cccc}
\toprule
  & AgileX-MV-Task1 & AgileX-MV-Task2 & AgileX-MV-Task3 & AgileX-MV-Task4\\
\midrule
$\pi_0$     & 0.1 & 0.0 & 0.0 & 0.1\\

$\pi_{0.5}$   & 0.3 & 0.0 & 0.3 & 0.1\\

XR-1     & 0.4 & 0.2 & 0.4 & 0.3\\
\midrule
\textbf{MIND-2}& \colorbox[HTML]{E0F4FF}{\textbf{0.5}} & \colorbox[HTML]{E0F4FF}{\textbf{0.8}} & \colorbox[HTML]{E0F4FF}{\textbf{0.4}} & \colorbox[HTML]{E0F4FF}{\textbf{0.7}} \\
\bottomrule
\end{tabular}
}
\end{table}

 \begin{figure}[h]
  \centering
\includegraphics[width=0.98\columnwidth]{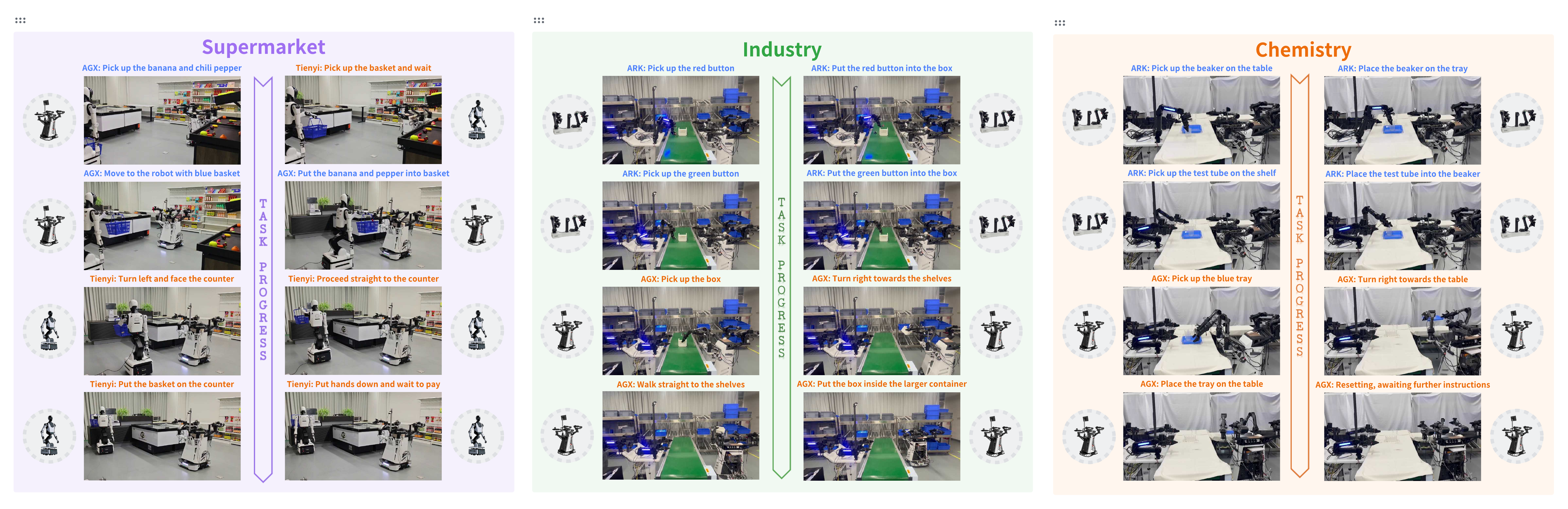}
  \caption{ Visualization of dual-robot collaborative tasks under three different environments.}
  \label{fig:dual}
\end{figure}
The MIND-2-VLM system can be viewed as a cloud-based robotic “brain" that determines the current stage of task execution. It is capable of controlling not only a single robot but also two robots with different morphologies. We design three collaborative tasks in which the Tian Yi and AgileX robots work together to accomplish challenging long-horizon mobile manipulation tasks across three distinct scenarios.
Figure~\ref{fig:dual} illustrates the dual-robot collaborative tasks we designed in three environments: supermarket scene, industry scene, and chemical laboratory scene.
In the supermarket, our task is to place shopping items in front of the checkout counter for payment. The AgileX robot picks up fruits from the fruit display and places them into the shopping basket held by Tian Yi, who then carries the basket to the checkout counter for settlement.
In the industrial scenario, a dual-arm robot places industrial switch components into a basket, after which an AgileX mobile robot (equipped with a mobile base) picks up the parts and places them onto an industrial guardrail.
In the chemical laboratory scene, a fixed dual-arm robot prepares a chemical solution, after which a mobile AgileX dual-arm robot transports the solution from the tray to the laboratory bench.

\begin{table}[htbp]
\centering
\caption{Success rates across three collaborative 
tasks. 
MIND-2 (Post Training)  is instantiated by fine-tuning InternVL3 and $\pi_{0.5}$ directly on data from the three multi-robot collaboration tasks.  
MIND-2 (Full-scale Training) is first pretrained on the full-scale mobile manipulation dataset using a fast-slow system architecture, and then further fine-tuned via post-training on data from three multi-robot collaboration tasks. MIND-2 (Offline RL), after full-scale training, we apply Implicit Q-Learning (IQL) to conduct offline reinforcement learning on the MIND-2-VLA.}
\label{tab:success_rate}
 \resizebox{0.95\columnwidth}{!}{
\begin{tabular}{l|c|c|c}
\toprule
      & \textbf{Supermarket Task} & \textbf{Industrial Task} &  \textbf{Chemical Laboratory Task} \\
\midrule
MIND-2 (Post Training) & 0.6    & 0.6    & 0.4    \\
MIND-2 (Full-scale Training) &0.8 &0.7 &0.4 \\
\textbf{MIND-2 (Offline RL)} & \colorbox[HTML]{E0F4FF}{\textbf{0.9}} & \colorbox[HTML]{E0F4FF}{\textbf{0.8}} & \colorbox[HTML]{E0F4FF}{\textbf{0.6}}\\
\bottomrule
\end{tabular}}
\end{table}
As shown in Table~\ref{tab:success_rate}, we evaluate three variants of MIND-2 on three multi-robot collaborative tasks: the Supermarket Task, the Industrial Task, and the Chemical Laboratory Task. 
MIND-2 (Post Training), which directly fine-tunes off-the-shelf VLA models (InternVL3 and $\pi_{0.5}$) on task-specific collaboration data, achieves moderate success rates (0.6, 0.6, and 0.4), indicating limited transfer without large-scale pretraining. 
In contrast, MIND-2 (Full-scale Training)—pretrained on the full-scale mobile manipulation dataset using a fast-slow architecture and then post-trained on collaboration data—shows consistent improvements, particularly in the Supermarket and Industrial tasks (0.8 and 0.7, respectively), though it still struggles with the complex sequencing required in the Chemical Laboratory Task (0.4). 
Notably, MIND-2 (Offline RL), which further applies Implicit Q-Learning (IQL) to the MIND-2-VLA policy after full-scale training, achieves the highest success rates across all three tasks (0.9, 0.8, and 0.6), demonstrating that offline reinforcement learning effectively refines action execution and enhances robustness in long-horizon, multi-agent scenarios.
MIND-2 models trained by Offline RL in Table~\ref{tab:success_rate} are 
trained on a dataset of 200 successful trajectories.
Then, we investigate whether incorporating failed trajectories into the IQL training framework can further improve the performance of MIND-2-VLA.  
Table~\ref{tab:failuresuccess} shows the performance of MIND-2 trained with IQL using varying ratios of successful to failed trajectories on three collaborative robot tasks.
We observe that the MIND-2-VLA model trained with IQL achieves the best performance when the number of failed trajectories matches that of successful ones.

\begin{table}[h]
\centering
\caption{Performance of MIND-2 trained with IQL under varying ratios of successful to failed trajectories on three collaborative robot tasks.}
\resizebox{0.85\columnwidth}{!}{
\begin{tabular}{c|c|c|c}
\toprule
\textbf{Success:Failure} & \textbf{Supermarket Task} & \textbf{Industrial Task} &  \textbf{Chemical Laboratory Task}  \\
\midrule
1:0 & 0.9&0.8 & 0.6    \\
2:1 & 1.0 & 0.9 & 0.6  \\
1:1 & \colorbox[HTML]{E0F4FF}{\textbf{1.0}} & \colorbox[HTML]{E0F4FF}{\textbf{1.0}} & \colorbox[HTML]{E0F4FF}{\textbf{0.8}}   \\
1:2 & 0.9 & 0.9 & 0.7  \\
\bottomrule
\end{tabular}
}
\label{tab:failuresuccess}
\end{table}

\subsection{Effectiveness of Tactile Sensing}

We select four mobile manipulation tasks from the AgileX platform to test the effectiveness of tactile sensing for imitation learning. 
We evaluate the impact of tactile sensing on imitation learning by selecting two representative VLA models, $\pi_{0.5}$ and XR-1, and integrating the collected tactile signals as part of the robot’s proprioceptive input during both training and evaluation. Specifically, tactile data is fused with other embodiment-aware observations (e.g., joint states, end-effector pose) to enrich the model’s perception of physical interaction, enabling more robust and fine-grained control in mobile manipulation tasks.

\begin{table}[htbp]
\centering
\caption{
  Performance comparison of $\pi_{0.5}$ and XR-1 with/without tactile sensor integration. \ding{51}/\ding{55} indicates the model was trained with/without tactile information.
}
\resizebox{0.95\columnwidth}{!}{
\begin{tabular}{l *{5}{c}}
\toprule
\textbf{Model} &\textbf{Tactile} & \textbf{AgileX-MV-Task1} & \textbf{AgileX-MV-Task2} & \textbf{AgileX-MV-Task3} & \textbf{AgileX-MV-Task4} \\
\midrule
$\pi_{0.5}$&\ding{55} & 0.3 & 0.0 & 0.3 & 0.1 \\
\cmidrule(lr){1-6}
$\pi_{0.5}$&\ding{51} & 0.4 & 0.1 & 0.5 & 0.2 \\
\cmidrule(lr){1-6}
XR-1&\ding{55} & 0.4 & 0.2 & 0.4 & 0.3 \\
\cmidrule(lr){1-6}
XR-1& \ding{51} &\colorbox[HTML]{E0F4FF}{\textbf{0.6}} & \colorbox[HTML]{E0F4FF}{\textbf{0.4}} & \colorbox[HTML]{E0F4FF}{\textbf{0.6}} & \colorbox[HTML]{E0F4FF}{\textbf{0.4}} \\
\bottomrule
\end{tabular}}
\label{tab:tensor_performance}
\vspace{-0.5em}
\end{table}

Based on Table~\ref{tab:tensor_performance}, we draw the following conclusions regarding the effectiveness of tactile sensing on the $\pi_{0.5}$ and XR-1 models.
Incorporating tactile information consistently improves the success rates across multiple mobile manipulation tasks, with particularly pronounced gains in scenarios requiring fine manipulation or physical interaction, demonstrating the complementary value of the tactile modality for perceiving physical interactions.
Tactile sensor provides more pronounced improvements for XR-1, especially in contact-rich, precision-demanding tasks, indicating its stronger capability in fusing multimodal proprioceptive signals. $\pi_{0.5}$ also benefits from tactile input, though to a lesser extent. These results underscore that tactile data, when integrated into embodiment-aware observations, complements vision and language and improves the robustness and generalization of VLA models in physically interactive environments.

\begin{figure*}[t]
  \centering
\subfloat{\includegraphics[width=0.95\textwidth]{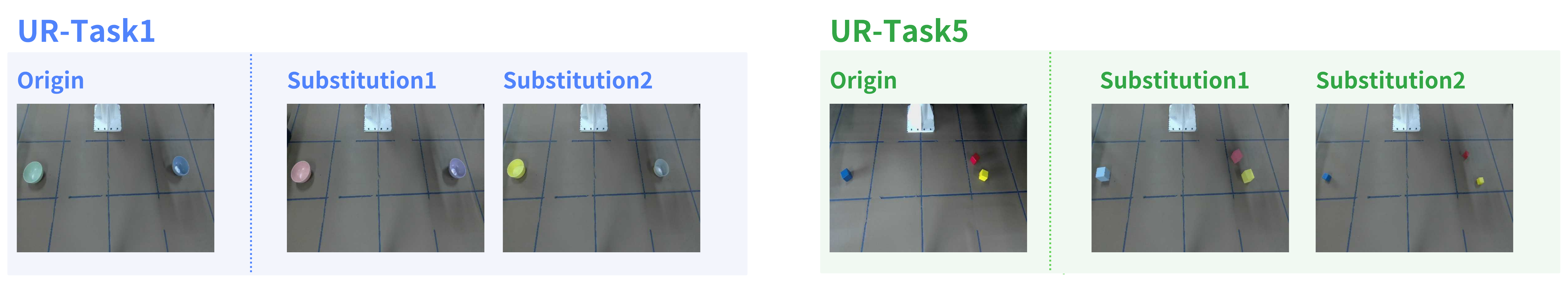}}
  \caption{
    Unseen objects used to evaluate the generalization ability of the VLA large models.
  }
  \label{fig:object_substitution}
\end{figure*}

\subsection{Object Substitution of VLA Models}

VLA models often emphasize strong generalization capabilities, yet ``generalization'' encompasses many dimensions, including task, environment, language instruction, and object variation. Among these, object-level generalization—the ability to manipulate novel or unseen physical objects in real-world settings—remains one of the most challenging.

To investigate this capability, we performed an experiment that uses the scale and diversity of our dataset. We train two representative VLA policies, $\pi_{0.5}$ and XR-1, on the two UR5e dual-arm manipulation tasks using real collected trajectories. At test time, we replace the original objects with functionally equivalent but visually or geometrically distinct alternatives (e.g., swapping a red button for a blue one, or using a different style of tray or container).
Especially, for UR-Task1, the manipulation object colors and shapes are changed (e.g., blue bowl → purple bowl, green bowl → pink bowl; or replaced with conical bowls), while keeping the same action structure. 
For UR-Task5,  we replace the original objects (e.g., wooden blocks) with foam or magnetic blocks of different materials (see Figure~\ref{fig:object_substitution}). 
This object substitution setting directly evaluates whether the policies can generalize beyond memorized object appearances and adapt to new instances—a critical requirement for practical deployment. Our results demonstrate that training on diverse, high-quality data significantly enhances robustness to object variation, underscoring the value of large-scale, multi-object datasets in advancing true operational generalization (see Table~\ref{tab:ur_generalization}).
\begin{table}[h]
\centering
\caption{Success rates on UR dual-arm tasks with object variations. UR-Task1 and UR-Task5 are base UR tasks; UR-Task1-a and UR-Task1-b modify objects in UR-Task1; UR-Task5-a and UR-Task5-b modify objects in UR-Task5.}
\label{tab:ur_generalization}
\begin{tabular}{lccc}
\toprule
\textbf{Task} & \textbf{Object Variation} & \textbf{$\pi_{0.5}$} & \textbf{XR-1} \\
\midrule
UR-Task1 & Original (e.g., blue/green bowls) & \colorbox[HTML]{E0F4FF}{\textbf{0.8}} & \colorbox[HTML]{E0F4FF}{\textbf{0.8}}\\
UR-Task1-a & Color/shape changed (e.g., purple/pink bowls) & \colorbox[HTML]{E0F4FF}{\textbf{0.8}} & \colorbox[HTML]{E0F4FF}{\textbf{0.8}}\\
UR-Task1-b & Geometry changed (e.g., conical bowls) & 0.6 & \colorbox[HTML]{E0F4FF}{\textbf{0.7}} \\
\cmidrule(lr){1-4}
UR-Task5 & Original (e.g., wooden blocks) & \colorbox[HTML]{E0F4FF}{\textbf{0.6}} &0.4 \\
UR-Task5-a & Material changed to foam blocks & \colorbox[HTML]{E0F4FF}{\textbf{0.6}}  &0.3\\
UR-Task5-b & Material changed to magnetic blocks & \colorbox[HTML]{E0F4FF}{\textbf{0.4}} &0.3 \\
\bottomrule
\end{tabular}
\end{table}

\subsection{Real and Simulation Experiments}
\label{exp:realsimulation}

To validate the effectiveness of simulation data in \ours, we conducted two sets of experiments. Our experimental evaluation focuses on three key aspects of simulation-based robot learning. First, we assess the intrinsic quality and behavioral plausibility of our simulated dataset by training and evaluating robot policies entirely within the digital twin of the Tien Kung dual-arm platform. Second, we investigate the utility of simulated data as a complementary resource to real-world demonstrations: specifically, we train imitation learning models (including ACT, Diffusion Policy, and XR-1) on hybrid datasets combining real and synthetic trajectories, and measure their performance on physical hardware. Third, we quantify the sim-to-real performance gap by comparing policy success rates under identical task specifications across simulation and the real robot. Together, these experiments provide a comprehensive validation of our simulation pipeline—not only as a high-fidelity data source in its own right, but also as an effective, low-cost augmentation strategy that enhances real-world policy robustness and generalization.






\subsubsection{Simulation Benchmark Tests}
To assess the quality and utility of our simulated dataset, we conduct experiments entirely within the digital twin of the Tien Kung dual-arm dexterous robot. We select four representative manipulation tasks and train single-task ACT, Diffusion Policy and XR-1 models using only our synthetic trajectories collected in simulation. 
The Tien Kung humanoid robot in four simulation tasks is  designed to assess bimanual coordination and object manipulation: (1) picking up a paper cup and placing it into a trash bin; (2) rotating a green pot handle from the 6 o’clock to the 9 o’clock position; (3) tidying a cluttered desktop by using both arms to place industrial switch objects into a basket; and (4) performing a coordinated sequence where the right arm pulls out a shelf, the left arm places a button onto it, and the right arm then pushes the shelf back in. These tasks vary in complexity, from single-arm actions to tightly coupled dual-arm operations, and involve diverse object interactions, making them well-suited for evaluating the generalization and robustness of vision-language-action policies in simulated humanoid settings.
Both training and evaluation are performed in the same simulated environment. As shown in Table~\ref{tab:simtest}, the consistent performance across tasks, which mirrors expected task difficulty and behavioral plausibility, demonstrates that our simulation pipeline generates coherent, high-fidelity data suitable for policy learning. This validates the realism and internal consistency of our simulated dataset, confirming its effectiveness as a scalable and low-cost resource for developing and benchmarking VLA-based manipulation policies.

\begin{table}[t]
\centering
\caption{Evaluation results on Tien Kung simulation tasks.}
\label{tab:simtest}
\begin{adjustbox}{max width=\textwidth}
\begin{tabular}{lcccc}
\toprule
\textbf{Method} & \textbf{Tien Kung-Task1} & \textbf{Tien Kung-Task2} & \textbf{Tien Kung-Task3} &\textbf{Tien Kung-Task4} \\
\midrule
\textbf{ACT} &43/50&34/50&33/50&18/50 \\
\textbf{Diffusion Policy }& 38/50&16/50&39/50
&22/50\\
\textbf{XR-1}& \colorbox[HTML]{E0F4FF}{\textbf{48/50}}&\colorbox[HTML]{E0F4FF}{\textbf{39/50}}&\colorbox[HTML]{E0F4FF}{\textbf{46/50}}
&\colorbox[HTML]{E0F4FF}{\textbf{31/50}} \\
\bottomrule
\end{tabular}
\end{adjustbox}
\end{table}

\begin{figure*}[t]
  \centering
\subfloat{\includegraphics[width=1.0\textwidth]{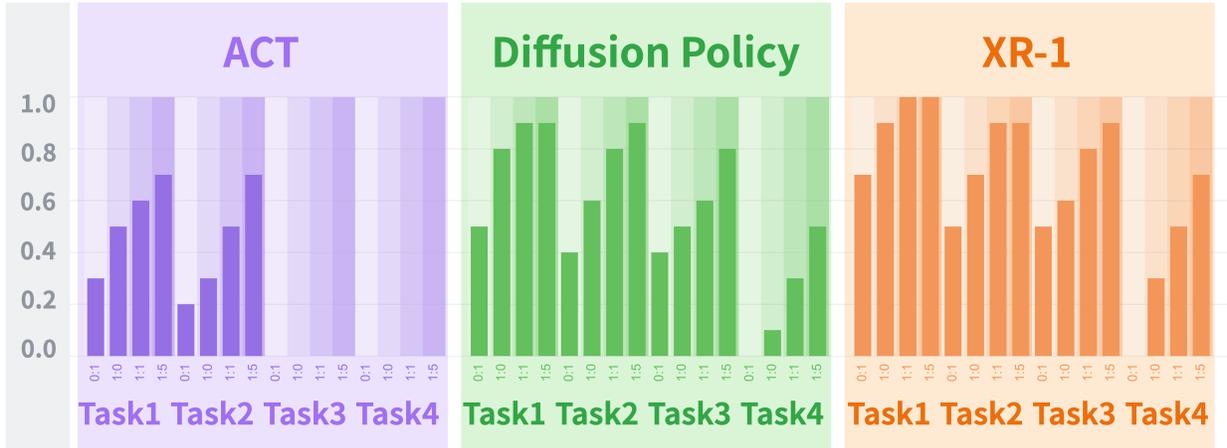}}
  \caption{
    Performance comparison of ACT, Diffusion Policy, and XR-1 across four distinct manipulation tasks under varying ratios of real to simulated training data (Real:Sim).
  }
  \label{fig:real_vs_sim}
\end{figure*}

\subsubsection{Co-training with Real and Simulation Data}
The evaluation is conducted on physical robots after training with mixed datasets combining real and simulated data at varying ratios. The rows correspond to different proportions of real-to-simulated data used during training:  0:1 (simulation only), 1:0 (real only), 1:1 (balanced), and 1:5 (more simulation added on top of the 1:1 setting).


Figure~\ref{fig:real_vs_sim} shows that all models benefit from co-training with real and simulated data, and importantly, further increasing the amount of simulation beyond a balanced mix (1:1$\rightarrow$1:5) can still improve real-robot performance. This trend is most evident on the more challenging tasks. For Diffusion Policy, adding more simulation on top of 1:1 increases success rates from 0.6 to 0.8 on Task~3 and from 0.3 to 0.5 on Task~4. XR-1 exhibits a similar pattern, improving from 0.8 to 0.9 on Task~3 and from 0.5 to 0.7 on Task~4 when moving from 1:1 to 1:5. These gains indicate that additional simulated experience can translate into better real-world execution, suggesting effective sim-to-real transfer rather than overfitting to simulation artifacts. In contrast, ACT improves on Task~1 and Task~2 but remains at 0.0 on Task~3 and Task~4 across all ratios, highlighting a clear limitation in handling contact-rich manipulation even with more simulated data.

Overall, the results demonstrate that scaling simulation data in mixed training continues to yield measurable real-robot improvements, particularly for XR-1 and Diffusion Policy on complex tasks, supporting the effectiveness of sim-to-real transfer in our setting.

\section{Discussion and Future Work}

In this work, we introduce \ours, a large-scale, multi-embodiment dataset specifically designed for bimanual robotic manipulation, featuring not only rich visual observations but also synchronized tactile feedback from high-resolution sensors.  
\ours includes demonstrations across six distinct dual-arm platforms ranging from fixed-base to mobile humanoids, comprising {\ntrajs} trajectories over {\ntasks} tasks, involving {\nobjs} objects and {\nskills} unique skills.  
All data are collected through an intelligent platform with rigorous quality assurance.  
By providing multimodal signals beyond vision alone, \ours enables research into perception-action loops that leverage touch for robust grasping, fine manipulation, and error recovery, which are essential for real-world deployment.


To rigorously evaluate the capabilities and limitations of modern robotic learning approaches, we position \ours as a comprehensive benchmark that supports both single-task imitation learning and multi-task VLA models. Notably, \ours contains a substantial volume of mobile manipulation data—collected across diverse real-world environments—enabling systematic evaluation in both mobile and stationary (fixed-base) settings, thereby offering a more holistic assessment of model generalization and robustness across heterogeneous robotic embodiments.

In addition, we perform three targeted ablation studies to dissect key factors in VLA-based manipulation: (1) object generalization, testing success on novel objects with unseen shapes, colors, or materials; (2) the role of tactile feedback, comparing policies trained with and without synchronized touch sensing to quantify its contribution to dexterity and robustness; and (3) sim-to-real transfer, evaluating whether pretraining on our high-fidelity photorealistic simulator—rigorously aligned with physical setups—improves real-world task execution. Together, these experiments not only validate the utility of \ours as a benchmark but also reveal critical insights into the data, architecture, and learning paradigms needed for scalable, embodied intelligence.

As an ongoing effort, we will continue expanding \ours with new embodiments, skills, and modalities (e.g., force-torque and audio) under the same rigorous collection protocol. We hope \ours not only serves as a ready-to-use resource but also inspires a shift toward systematic, quality-first data curation in embodied AI.


\section*{Acknowledgments}

This dataset and benchmark for robotic arm manipulation tasks represent a significant systems engineering undertaking, made possible only through extensive collaboration among researchers across multiple disciplines.

The successful development of this work would not have been achievable without the unwavering dedication, diverse expertise, and sustained contributions of numerous individuals throughout all phases of the project.
We extend our deepest gratitude to the following colleagues (in alphabetical order) for their invaluable support:
Congjia Su, Dapeng Wang, Guang Yang, Guangyu Li, Huijuan Ma, Jian Xiao, Jianwei Guo, Jianwei Sun, Jianyu Dong, Jiaxing Wei, Jieyu Zhang, Kun Niu, Mingxuan Guo, Panpan Chen, Peng Guo, Pengwei Zhang, Qiang Zhang, Qichun Liu, Qiu Cui, Rongwei Ren, Shiwei Jiao, Shuang Wang, Shuguang Qiao, Weixin Zhang, Wenjun Ren, Xiangquan Gao, Xiaozhu Ju, Yang Pan, Yanhui Ma, Yaning Hu, Yaowen Xu, Yinuo Zhao, Yulin Luo, and Zhifei Xiang.

We also sincerely thank the many additional contributors who played essential roles in data collection, quality assurance, annotation, and testing. Their collective efforts and technical insights were indispensable to the realization of this research.
This collaborative endeavor reflects a shared commitment to advancing the field of robotic manipulation, and we are profoundly grateful for everyone’s contributions.

This work is supported by the National Natural Science Foundation of China (62476011), and by the Beijing Natural Science Foundation (L252060).

\section*{Author Contributions}

\begin{itemize}
  \item \textbf{Project Lead:}
    Zhengping Che
    
  \item \textbf{Project Coordinators:}
    Chengkai Hou, Jiaming Liu, and Kun Wu

  \item \textbf{Data Collection and Processing:}
    Guangrun Li, Jingyang He, Chengkai Hou, Di Wu, Kun Wu, Xinhua Wang, Shichao Fan, Meng Li, Zhen Zhao, Ning Liu, Yuxue Zhang, Zhiyuan Xu, Pei Ren, Junjie Ji, Nuowei Han, and Xiangju Mi

  \item \textbf{Dataset Annotation:}
    Di Wu and Qiuxuan Feng

  \item \textbf{Algorithm Training and Testing:}
    \begin{itemize}
      \item MIND-2: Fei Liao, Qiuxuan Feng, and Chengkai Hou
      \item ACT: Jingyang He and Guangrun Li
      \item 3D Diffusion Policy (DP3): Yankai Fu, Jingyang He, and Guangrun Li
      \item Dense Policy: Chengkai Hou
      \item UVA: Yaoxu Lv and Guangrun Li
      \item $\pi_0$ /$\pi_{0.5}$: Chenyang Gu, Jiaming Liu, and Zhuoyang Liu
      \item HybridVLA: Jiaming Liu and Zhuoyang Liu
      \item XR-1: Shichao Fan, Kun Wu, and Xinhua Wang 
      \item Simulation Benchmark Framework: Zhao Jin, Tao Li, Yuheng Zhang, and Kun Wu
    \end{itemize}

  \item \textbf{System and Infrastructure Development:}
    \begin{itemize}
      \item Data Collection System: Zhiyuan Xu, Pei Ren, Junjie Ji, and Yixue Zhang
      \item Mobile Manipulation Pipeline: Fei Liao and Guangrun Li
      \item Tactile Integration and Ablation: Guangrun Li, Langzhe Gu, and Chengkai Hou
    \end{itemize}

  \item \textbf{Paper Writing:}
    Chengkai Hou, Kun Wu, Gaole Dai, and Zhengping Che

  \item \textbf{Project Support:}
    Haonan Liu, Jiaming Liu, Yaoxu Lü, Xiangju Mi, Nuowei Han, Zhuoyang Liu, Chenyang Gu, Yankai Fu, Zhao Jin, Tao Li, and Langzhe Gu

  \item \textbf{Project Advisors:}
    Jian Tang, Shanghang Zhang, and Kuan Cheng

  \item \textbf{Corresponding Authors:}
    Jian Tang and Shanghang Zhang
\end{itemize}



%
\bibliography{references} 
\bibliographystyle{sciencemag}



%
%
%
%
%
%




\end{document}